\documentclass[10pt,twocolumn,letterpaper]{article}

\usepackage{iccv}
\usepackage{times}
\usepackage{epsfig}
\usepackage{graphicx}
\usepackage{amsmath}
\usepackage{amssymb}
\usepackage{caption}
\usepackage{color}
\usepackage{subfig}
\usepackage{lipsum}

\newcommand\blfootnote[1]{%
  \begingroup
  \renewcommand\thefootnote{}\footnote{#1}%
  \addtocounter{footnote}{-1}%
  \endgroup
}

\usepackage[pagebackref=true,breaklinks=true,letterpaper=true,colorlinks,bookmarks=false]{hyperref}

\newcommand{\reffig}[1]{Figure~\ref{fig:#1}}
\newcommand{\refsec}[1]{Section~\ref{sec:#1}}

\newcommand{\reftbl}[1]{Table~\ref{tbl:#1}}

\newcommand{\refeq}[1]{Equation~\ref{eq:#1}}

\newcommand{\lblfig}[1]{\label{fig:#1}}
\newcommand{\lblsec}[1]{\label{sec:#1}}
\newcommand{\lbleq}[1]{\label{eq:#1}}
\newcommand{\lbltbl}[1]{\label{tbl:#1}}

\newcommand{\shortcite}[1]{\cite{#1}}
\newcommand{\ignorethis}[1]{}
\iccvfinalcopy %

\ificcvfinal\fi

\newcommand{\norm}[1]{\lVert#1\rVert}

\begin{document}

\title{Unpaired Image-to-Image Translation \\ using Cycle-Consistent Adversarial Networks}

\author{
Jun-Yan Zhu\thanks{empty}
\and
Taesung Park\footnotemark[1]
\and 
Phillip Isola
\and
Alexei A. Efros
\and
Berkeley AI Research (BAIR) laboratory, UC Berkeley
}

\twocolumn[{%
\renewcommand\twocolumn[1][]{#1}%
\vspace{-3em}
\maketitle
\vspace{-3em}
\begin{center}
    \centering
    \includegraphics[width=\linewidth]{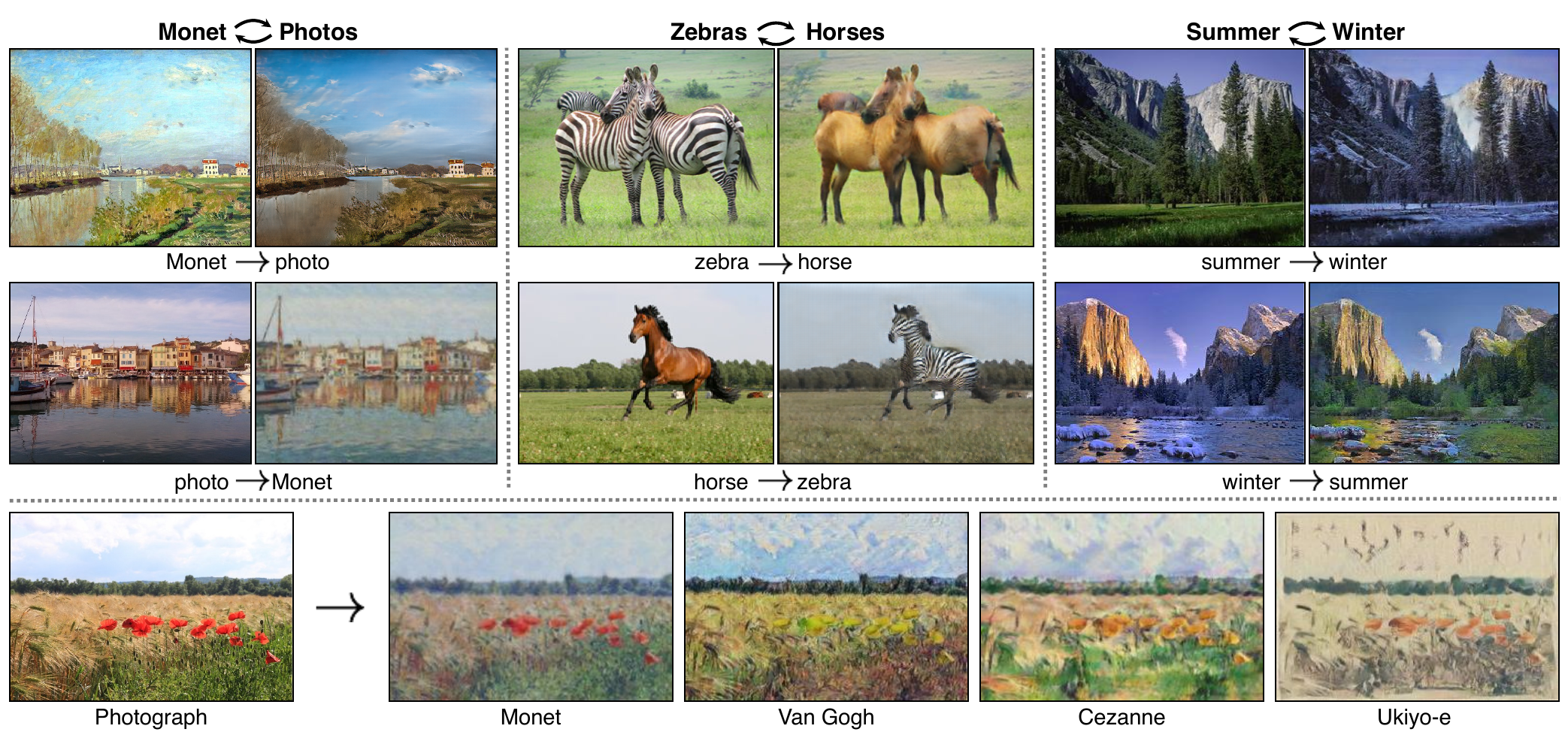}
    \vspace{-2.5em}
    \captionof{figure}{
    Given any two unordered image collections $X$ and $Y$, our algorithm learns to automatically ``translate'' an image from one into the other and vice versa: {\em (left)} Monet paintings and landscape photos from Flickr; {\em (center)} zebras and horses from ImageNet; {\em (right)} summer and winter Yosemite photos from Flickr.  Example application {\em (bottom)}: using a collection of paintings of famous artists, our method learns to render natural photographs into the respective styles.
    }
    \lblfig{teaser}
\end{center}%
}]

\begin{abstract}
\vspace{-1.5em}
Image-to-image translation is a class of vision and graphics problems where the goal is to learn the mapping between an input image and an output image using a training set of aligned image pairs.   
However, for many tasks, paired training data will not be available.  We present an approach for learning to translate an image from a source domain $X$ to a target domain $Y$ in the absence of paired examples.  Our goal is to learn a mapping $G: X \rightarrow Y$ such that the distribution of images from $G(X)$ is indistinguishable from the distribution $Y$ using an adversarial loss.  
Because this mapping is highly under-constrained, we couple it with an inverse mapping $F: Y \rightarrow X$ and introduce a {\em cycle consistency loss} 
to enforce $F(G(X)) \approx X$ (and vice versa).
Qualitative results are presented on several tasks where paired training data does not exist, including collection style transfer, object transfiguration, season transfer, photo enhancement, etc.  Quantitative comparisons against several prior methods demonstrate the superiority of our approach.
\end{abstract}

\section{Introduction}
\lblsec{intro}

What did Claude Monet see as he placed his easel by the bank of the Seine near Argenteuil on a lovely spring day in 1873 (\reffig{teaser}, top-left)? A color photograph, had it been invented, may have documented a crisp blue sky and a glassy river reflecting it. Monet conveyed his {\em impression} of this same scene through wispy brush strokes and a bright palette.\blfootnote{* indicates equal contribution}

What if Monet had happened upon the little harbor in Cassis on a cool summer evening (\reffig{teaser}, bottom-left)? A brief stroll through a gallery of Monet paintings makes it possible to imagine how he would have rendered the scene: perhaps in pastel shades, with abrupt dabs of paint, and a somewhat flattened dynamic range.

\begin{figure}
 \centering
 \includegraphics[width=1.0\hsize]{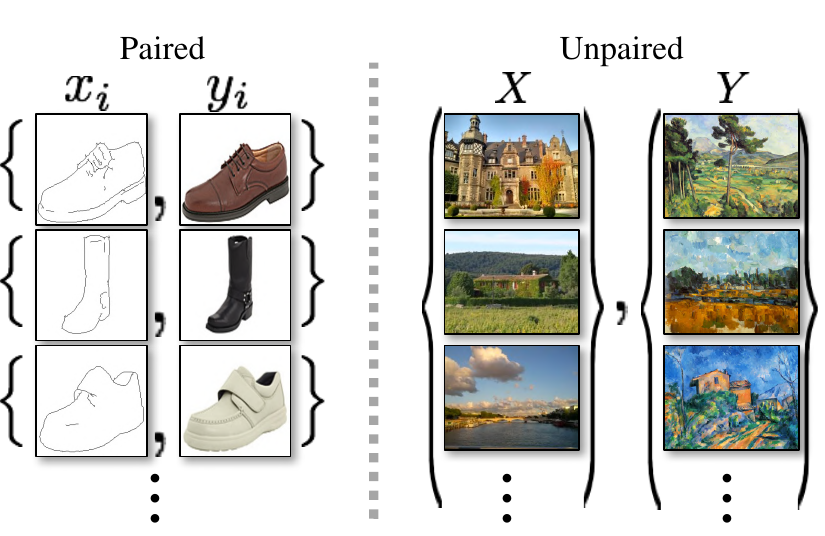}
 \vspace{-0.3in}
  \caption{\emph{Paired} training data (left) consists of training examples $\{x_i,y_i\}_{i=1}^N$, where the correspondence between $x_i$ and $y_i$ exists~\cite{isola2016image}. We instead consider \emph{unpaired} training data (right), consisting of a source set $\{x_i\}_{i=1}^N$ ($x_i \in X$) and a target set $\{y_j\}_{j=1}^M$ ($y_j \in Y$), with no information provided as to which $x_i$ matches which $y_j$.}
 \lblfig{paired_unpaired}
 \vspace{-0.2in}
\end{figure}

We can imagine all this despite never having seen a side by side example of a Monet painting next to a photo of the scene he painted. Instead, we have knowledge of the set of Monet paintings and of the set of landscape photographs. We can reason about the stylistic differences between these two sets, and thereby imagine what a scene might look like if we were to ``translate'' it from one set into the other.

In this paper, we present a method that can learn to do the same: capturing special characteristics of one image collection and figuring out how these characteristics could be translated into the other image collection, all in the absence of any paired training examples. 

This problem can be more broadly described as image-to-image translation~\cite{isola2016image}, converting an image from one representation of a given scene, $x$, to another, $y$, e.g., grayscale to color, image to semantic labels, edge-map to photograph.
Years of research in computer vision, image processing, computational photography, and graphics have produced powerful translation systems in the supervised setting, where example image pairs $\{x_i,y_i\}_{i=1}^N$ are available (\reffig{paired_unpaired}, left), e.g., 
\cite{eigen2015predicting, hertzmann2001image, isola2016image, johnson2016perceptual, laffont2014transient, long2015fully, shih2013data, wang2016generative, xie2015holistically, zhang2016colorful}.
However, obtaining paired training data can be difficult and expensive. For example, only a couple of datasets exist for tasks like semantic segmentation (e.g.,~\cite{Cordts2016Cityscapes}), and they are relatively small.  
Obtaining input-output pairs for graphics tasks like artistic stylization can be even more difficult since the desired output is highly complex, typically requiring artistic authoring. For many tasks, like object transfiguration (e.g., zebra$\leftrightarrow$horse, ~\reffig{teaser} top-middle), the desired output is not even well-defined.

We therefore seek an algorithm that can learn to translate between domains without paired input-output examples (\reffig{paired_unpaired}, right). We assume there is some underlying relationship between the domains -- for example, that they are two different renderings of the same underlying scene -- and seek to learn that relationship. Although we lack supervision in the form of paired examples, we can exploit supervision at the level of sets: we are given one set of images in domain $X$ and a different set in domain $Y$. We may train a mapping $G: X \rightarrow Y$ such that the output $\hat{y} = G(x)$, $x \in X$, is indistinguishable from images $y \in Y$ by an adversary trained to classify $\hat{y}$ apart from $y$. In theory, this objective can induce an output distribution over $\hat{y}$ that matches the empirical distribution $p_{data}(y)$ (in general, this requires $G$ to be stochastic)~\cite{goodfellow2014generative}. The optimal $G$ thereby translates the domain $X$ to a domain $\hat{Y}$ distributed identically to $Y$. However, such a translation does not guarantee that an individual input $x$ and output $y$ are paired up in a meaningful way -- there are infinitely many mappings $G$ that will induce the same distribution over $\hat{y}$. Moreover, in practice, we have found it difficult to optimize the adversarial objective in isolation: standard procedures often lead to the well-known problem of mode collapse, where all input images map to the same output image and the optimization fails to make progress~\cite{goodfellow2016nips}.

These issues call for adding more structure to our objective. Therefore, we exploit the property that translation should be ``cycle consistent", in the sense that if we translate, e.g., a sentence from English to French, and then translate it back from French to English, we should arrive back at the original sentence~\cite{brislin1970back}. Mathematically, if we have a translator $G: X \rightarrow Y$ and another translator $F: Y \rightarrow X$, then $G$ and $F$ should be inverses of each other, and both mappings should be bijections. We apply this structural assumption by training both the mapping $G$ and $F$ simultaneously, and adding a \emph{cycle consistency loss}~\cite{zhou2016learning} that encourages $F(G(x)) \approx x$ and $G(F(y)) \approx y$. Combining this loss with adversarial losses on domains $X$ and $Y$ yields our full objective for unpaired image-to-image translation.

We apply our method to a wide range of applications, including collection style transfer, object transfiguration, season transfer and photo enhancement. We also compare against previous approaches that rely either on hand-defined factorizations of style and content, or on shared embedding functions, and show that our method outperforms these baselines. We provide both \href{https://github.com/junyanz/pytorch-CycleGAN-and-pix2pix}{PyTorch} and \href{https://github.com/junyanz/CycleGAN}{Torch} implementations. Check out more results at our \href{https://junyanz.github.io/CycleGAN/}{website}.

\begin{figure*}[th]
 \centering
 \includegraphics[width=1.0\hsize]{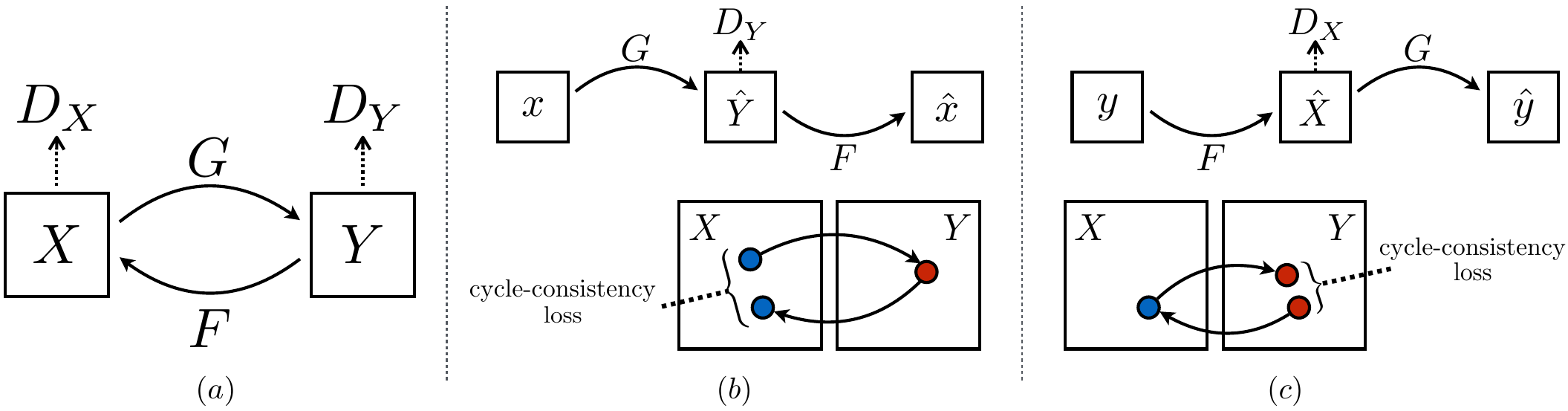}
 \vspace{-0.2in}
  \caption{(a) Our model contains two mapping functions $G: X \rightarrow Y$ and $F: Y \rightarrow X$, and associated adversarial discriminators $D_Y$ and $D_X$. $D_Y$ encourages $G$ to translate $X$ into outputs indistinguishable from domain $Y$, and vice versa for $D_X$ and $F$. To further regularize the mappings, we introduce two {\it cycle consistency losses} that capture the intuition that if we translate from one domain to the other and back again we should arrive at where we started: (b) forward cycle-consistency loss: $x\rightarrow G(x) \rightarrow F(G(x)) \approx x$, and (c) backward cycle-consistency loss: $y \rightarrow F(y) \rightarrow G(F(y)) \approx y$ }
 \lblfig{overview}
 \vspace{-0.2in}
\end{figure*}

\section{Related work}
\lblsec{relatedwork}

{\bf Generative Adversarial Networks  (GANs)}~\cite{goodfellow2014generative,zhao2016energy} have achieved impressive results in image generation~\cite{denton2015deep,radford2015unsupervised}, image editing~\cite{zhu2016generative}, and representation learning~\cite{radford2015unsupervised,salimans2016improved,mathieu2016disentangling}.　
Recent methods adopt the same idea for conditional image generation applications, such as text2image~\cite{reed2016generative}, image inpainting~\cite{pathak2016context}, and future prediction~\cite{mathieu2015deep}, as well as to other domains like videos~\cite{vondrick2016generating} and 3D data~\cite{wu2016learning}. 
The key to GANs' success is the idea of an \textit{adversarial loss} that forces the generated images to be, in principle, indistinguishable from real photos.  
This loss is particularly powerful for image generation tasks,  
as this is exactly the objective that much of computer graphics aims to optimize.  We adopt an adversarial loss to learn the mapping such that the translated images cannot be distinguished from images in the target domain.

{\bf Image-to-Image Translation}
The idea of image-to-image translation goes back at least to Hertzmann et al.'s Image Analogies~\cite{hertzmann2001image}, who employ a non-parametric texture model~\cite{efros1999texture} on a single input-output training image pair.
More recent approaches use a \emph{dataset} of input-output examples to learn a parametric translation function using CNNs (e.g.,~\cite{long2015fully}). Our approach builds on the ``pix2pix" framework of Isola et al.~\cite{isola2016image}, which uses a conditional generative adversarial network~\cite{goodfellow2014generative} to learn a mapping from input to output images. Similar ideas have been applied to various tasks such as generating photographs from sketches~\cite{sangkloy2016scribbler} or from attribute and semantic layouts~\cite{karacan2016learning}. However, unlike the above prior work, we learn the mapping without paired training examples. 

{\bf Unpaired Image-to-Image Translation}
Several other methods also tackle the unpaired setting, where the goal is to relate two data domains: $X$ and $Y$. Rosales et al.~\shortcite{rosales2003unsupervised} propose a Bayesian framework that includes a prior based on a patch-based Markov random field computed from a source image and a likelihood term obtained from multiple style images. More recently, CoGAN~\cite{liu2016coupled} and cross-modal scene networks~\cite{aytar2016cross} use a weight-sharing strategy to learn a common representation across domains. 
Concurrent to our method, Liu et al.~\shortcite{liu2017unsupervised} extends the above framework with a combination of variational autoencoders~\cite{kingma2013auto} and generative adversarial networks~\cite{goodfellow2014generative}.
Another line of concurrent work~\cite{shrivastava2016learning,taigman2016unsupervised,bousmalis2016unsupervised} encourages the input and output to share specific ``content" features even though they may differ in ``style``. These methods also use adversarial networks, with additional terms to enforce the output to be close to the input in a predefined metric space, such as class label space~\cite{bousmalis2016unsupervised}, image pixel space~\cite{shrivastava2016learning}, and image feature space~\cite{taigman2016unsupervised}. 

Unlike the above approaches, our formulation does not rely on any task-specific, predefined similarity function between the input and output, nor do we assume that the input and output have to lie in the same low-dimensional embedding space. This makes our method a general-purpose solution for many vision and graphics tasks. We directly compare against several prior and contemporary approaches in \refsec{comparison}.

{\bf Cycle Consistency}
The idea of using transitivity as a way to regularize structured data has a long history.  In visual tracking, enforcing simple forward-backward consistency has been a standard trick for decades~\cite{kalal2010forward,sundaram2010dense}. In the language domain, verifying and improving translations via ``back translation and reconciliation'' is a technique used by human translators~\cite{brislin1970back} (including, humorously, by Mark Twain~\cite{twain1903}), as well as by machines~\cite{he2016dual}. 
More recently, higher-order cycle consistency has been used in
structure from motion~\cite{zach2010disambiguating},
3D shape matching~\cite{huang2013consistent}, co-segmentation~\cite{wang2013image}, dense semantic alignment~\cite{zhou2015flowweb,zhou2016learning}, and depth estimation~\cite{godard2016unsupervised}.  Of these, Zhou et al.~\cite{zhou2016learning} and Godard et al.~\cite{godard2016unsupervised} are most similar to our work, as they use a {\em cycle consistency loss} as a way of using transitivity to supervise CNN training.
In this work, we are introducing a similar loss to push $G$ and $F$ to be consistent with each other.  Concurrent with our work, in these same proceedings, Yi et al.~\cite{yi2017dualgan} independently use a similar objective for unpaired image-to-image translation, inspired by dual learning in machine translation~\cite{he2016dual}.

{\bf Neural Style Transfer}~\cite{gatys2015neural,johnson2016perceptual,ulyanov2016texture,gatys2016preserving} is another way to perform image-to-image translation, which synthesizes a novel image by combining the content of one image with the style of another image (typically a painting) based on matching the Gram matrix statistics of pre-trained deep features.  Our primary focus, on the other hand, is learning the mapping between two image collections, rather than between two specific images, by trying to capture correspondences between higher-level appearance structures. Therefore, our method can be applied to other tasks, such as painting$\rightarrow$ photo, object transfiguration, etc. where single sample transfer methods do not perform well. We compare these two methods in ~\refsec{application}.

\section{Formulation}
\lblsec{method}

Our goal is to learn mapping functions between two domains $X$ and $Y$ given training samples $\{x_i\}_{i=1}^N$ where $x_i \in X$ and $\{y_j\}_{j=1}^M$ where $y_j \in Y$\footnote{We often omit the subscript $i$ and $j$ for simplicity.}. We denote the data distribution as $x\sim p_{data}(x)$ and $y\sim p_{data}(y)$. As illustrated in \reffig{overview} (a), our model includes two mappings $G: X\rightarrow Y$ and $F: Y\rightarrow X$. In addition, we introduce two adversarial discriminators $D_X$ and $D_Y$, where $D_X$ aims to distinguish between images $\{x\}$ and translated images $\{F(y)\}$; in the same way, $D_Y$ aims to discriminate between $\{y\}$ and $\{G(x)\}$. Our objective contains two types of terms: \textit{adversarial losses}~\cite{goodfellow2014generative} for matching the distribution of generated images to the data distribution in the target domain; and \textit{cycle consistency losses} to prevent the learned mappings $G$ and $F$ from contradicting each other.

\subsection{Adversarial Loss}
We apply adversarial losses~\cite{goodfellow2014generative} to both mapping functions. For the mapping function $G: X\rightarrow Y$ and its discriminator $D_Y$, we express the objective as: 
\begin{align}
    \mathcal{L}_{\text{GAN}}(G,D_Y,X,Y) =& \ \mathbb{E}_{y \sim p_{\text{data}}(y)}[\log D_Y(y)] \nonumber \\
   +& \ \mathbb{E}_{x \sim p_{\text{data}}(x)}[\log (1-D_Y(G(x))],\lbleq{GAN}
\end{align}
where $G$ tries to generate images $G(x)$ that look similar to images from domain $Y$, while $D_Y$ aims to distinguish between translated samples $G(x)$ and real samples $y$. $G$ aims to minimize this objective against an adversary $D$ that tries to maximize it, i.e., $\min_G \max_{D_Y} \mathcal{L}_{\text{GAN}}(G,D_Y,X,Y)$.
We introduce a similar adversarial loss for the mapping function $F:Y\rightarrow X$ and its discriminator $D_X$ as well: i.e., $\min_F \max_{D_X} \mathcal{L}_{\text{GAN}}(F,D_X,Y,X)$.

\begin{figure}[t]
\begin{center}
\includegraphics[width=1.0\linewidth]{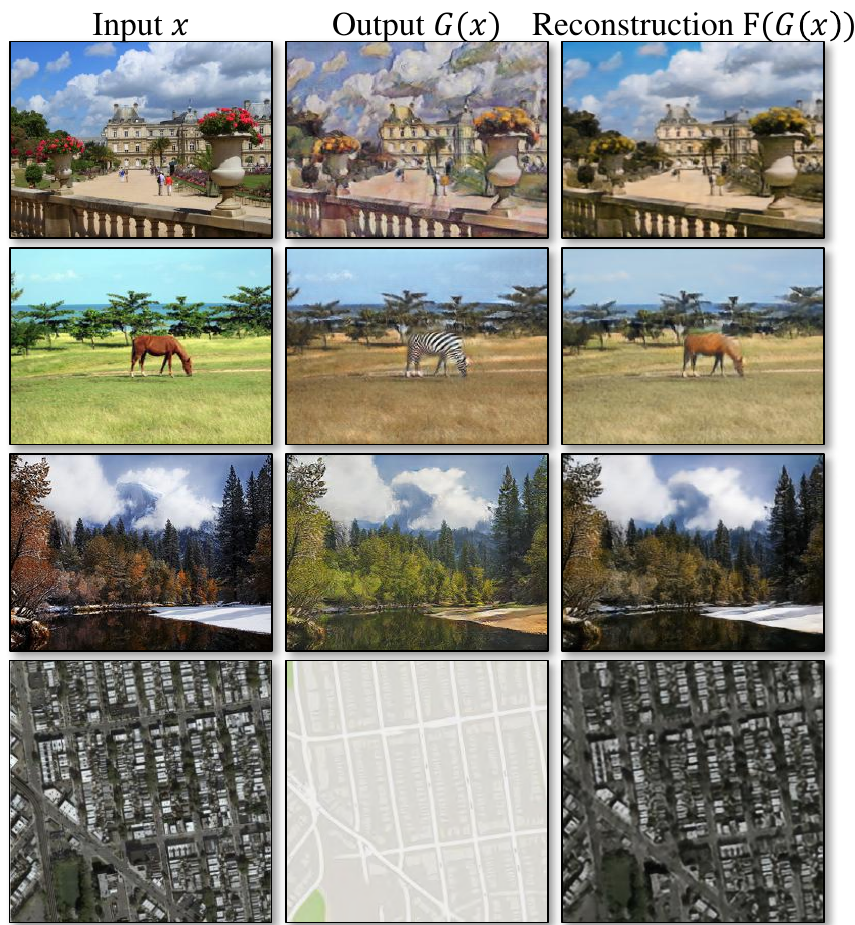}
\end{center}
 \vspace{-5 mm}
 \caption{The input images $x$, output images $G(x)$ and the reconstructed images $F(G(x))$ from various experiments. From top to bottom: photo$\leftrightarrow$Cezanne, horses$\leftrightarrow$zebras,  winter$\rightarrow$summer Yosemite, aerial photos$\leftrightarrow$Google maps.}
  \vspace{-5 mm}
\lblfig{reconstruction}
\end{figure}

\subsection{Cycle Consistency Loss}
Adversarial training can, in theory, learn mappings $G$ and $F$ that produce outputs identically distributed as target domains $Y$ and $X$ respectively (strictly speaking, this requires $G$ and $F$ to be stochastic functions)~\cite{goodfellow2016nips}. However, with large enough capacity, a network can map the same set of input images to any random permutation of images in the target domain, where any of the learned mappings can induce an output distribution that matches the target distribution. Thus, adversarial losses alone cannot guarantee that the learned function can map an individual input $x_i$ to a desired output $y_i$. To further reduce the space of possible mapping functions, we argue that the learned mapping functions should be cycle-consistent: as shown in \reffig{overview} (b), for each image $x$ from domain $X$, the image translation cycle should be able to bring $x$ back to the original image, i.e., $x \rightarrow G(x) \rightarrow F(G(x)) \approx x$. We call this {\em forward cycle consistency}. Similarly, as illustrated in \reffig{overview} (c), for each image $y$ from domain $Y$, $G$ and $F$ should also satisfy {\em backward cycle consistency}: $y \rightarrow F(y) \rightarrow G(F(y)) \approx y$.
We incentivize this behavior using a \emph{cycle consistency loss}:
\begin{align}
    \mathcal{L}_{\text{cyc}}(G, F) =  & \ \mathbb{E}_{x\sim p_{\text{data}}(x)}[\norm{F(G(x))-x}_1] \nonumber \\ 
    + &\ \mathbb{E}_{y\sim p_{\text{data}}(y)}[\norm{G(F(y))-y}_1].\lbleq{cycle}
\end{align}
In preliminary experiments, we also tried replacing the L1 norm in this loss with an adversarial loss between $F(G(x))$ and $x$, and between $G(F(y))$ and $y$, but did not observe improved performance.

The behavior induced by the cycle consistency loss can be observed in \reffig{reconstruction}: the reconstructed images $F(G(x))$ end up matching closely to the input images $x$.

\bigskip
\subsection{Full Objective}
Our full objective is:
\begin{align}
     \mathcal{L}(G,F,D_X,D_Y) = &、 \mathcal{L}_{\text{GAN}}(G,D_Y,X,Y) \nonumber \\
    +&\ \mathcal{L}_{\text{GAN}}(F,D_X,Y,X) \nonumber \\
    +& \  \lambda \mathcal{L}_{\text{cyc}}(G, F),\lbleq{full_objective}
\end{align}
where $\lambda$ controls the relative importance of the two objectives. We aim to solve:
\begin{equation}
    G^*,F^* = \arg\min_{G,F}\max_{D_x,D_Y} \mathcal{L}(G, F, D_X, D_Y).
    \lbleq{minmax}
\end{equation}

Notice that our model can be viewed as training two ``autoencoders"~\cite{hinton2006reducing}: we learn one autoencoder $F \circ G: X \rightarrow X$ jointly with another $G \circ F: Y \rightarrow Y$. However, these autoencoders each have special internal structures: they map an image to itself via an intermediate representation that is a translation of the image into another domain. Such a setup can also be seen as a special case of ``adversarial autoencoders"~\cite{makhzani2015adversarial}, which use an adversarial loss to train the bottleneck layer of an autoencoder to match an arbitrary target distribution. In our case, the target distribution for the $X \rightarrow X$ autoencoder is that of the domain $Y$.

In \refsec{ablation}, we compare our method against ablations of the full objective, including the adversarial loss $\mathcal{L}_{\text{GAN}}$ alone and the cycle consistency loss $\mathcal{L}_{\text{cyc}}$ alone, and empirically show that both objectives play critical roles in arriving at high-quality results. We also evaluate our method with only cycle loss in one direction and show that a single cycle is not sufficient to regularize the training for this under-constrained problem.

\section{Implementation}
\lblsec{implementation}

\paragraph{Network Architecture}
We adopt the architecture for our generative networks from Johnson et al.~\shortcite{johnson2016perceptual} who have shown impressive results for neural style transfer and super-resolution. This network contains three convolutions, several residual blocks~\cite{he2016deep},  two fractionally-strided convolutions with stride $\frac{1}{2}$,  and one convolution that maps features to RGB. We use $6$ blocks for $128\times128$ images and $9$ blocks for $256\times 256$ and higher-resolution training images. Similar to Johnson et al.~\shortcite{johnson2016perceptual}, we use instance normalization~\cite{ulyanov2016instance}. For the discriminator networks we use $70\times 70$ PatchGANs~\cite{isola2016image,li2016precomputed,ledig2016photo}, which aim to classify whether $70 \times 70$ overlapping image patches are real or fake. Such a patch-level discriminator architecture has fewer parameters than a full-image discriminator and can work on arbitrarily-sized images in a fully convolutional fashion~\cite{isola2016image}.

\paragraph{Training details}
We apply two techniques from recent works to stabilize our model training procedure. First, for $\mathcal{L}_{\text{GAN}}$ (\refeq{GAN}), we replace the negative log likelihood objective by a least-squares loss~\cite{mao2017least}. This loss is more stable during training and generates higher quality results. In particular, for a GAN loss $\mathcal{L}_{\text{GAN}}(G,D,X,Y)$, we train the $G$ to minimize $\mathbb{E}_{x \sim p_{\text{data}}(x)}[(D(G(x))-1)^2]$ and train the $D$ to minimize $\mathbb{E}_{y \sim p_{\text{data}}(y)}[(D(y)-1)^2]+\mathbb{E}_{x \sim p_{\text{data}}(x)}[D(G(x))^2]$.

Second, to reduce model oscillation~\cite{goodfellow2016nips}, we follow Shrivastava et al.'s strategy~\cite{shrivastava2016learning} and update the discriminators using a history of generated images rather than the ones produced by the
latest generators. We keep an image buffer that stores the $50$ previously created images.

For all the experiments, we set $\lambda=10$ in \refeq{full_objective}.
We use the Adam solver~\cite{kingma2014adam} with a batch size of $1$. All networks were trained from scratch with a learning rate of $0.0002$. We keep the same learning rate for the first $100$ epochs and linearly decay the rate to zero over the next $100$ epochs.
Please see the appendix (\refsec{appendix}) for more details about the datasets, architectures, and training procedures.

\section{Results}
\lblsec{results}
We first compare our approach against recent methods for unpaired image-to-image translation on paired datasets where ground truth input-output pairs are available for evaluation. We then study the importance of both the adversarial loss and the cycle consistency loss and compare our full method against several variants. Finally, we demonstrate the generality of our algorithm on a wide range of applications where paired data does not exist. For brevity, we refer to our method as {\tt CycleGAN}. The \href{https://github.com/junyanz/pytorch-CycleGAN-and-pix2pix}{PyTorch} and \href{https://github.com/junyanz/CycleGAN}{Torch} code, models, and full results can be found at our \href{https://junyanz.github.io/CycleGAN/}{website}.

\subsection{Evaluation}
\lblsec{comparison}

\begin{figure*}[t]
\begin{center}
\includegraphics[width=0.95\linewidth]{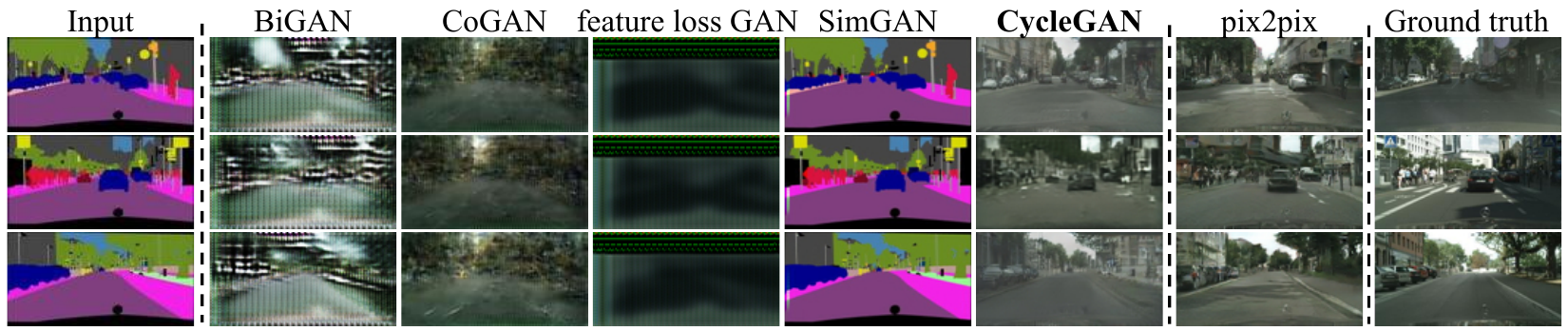}
\end{center}
 \vspace{-4 mm}
 \caption{Different methods for mapping labels$\leftrightarrow$photos trained on Cityscapes images. From left to right: input, BiGAN/ALI~\cite{donahue2016adversarial,dumoulin2016adversarially}, CoGAN~\cite{liu2016coupled},  feature loss + GAN, SimGAN~\cite{shrivastava2016learning}, CycleGAN (ours), pix2pix~\cite{isola2016image} trained on paired data, and ground truth.}
  \vspace{-4 mm}
\lblfig{cityscapes_baseline}
\end{figure*}

\begin{figure*}[t]
\begin{center}
\includegraphics[width=0.95\linewidth]{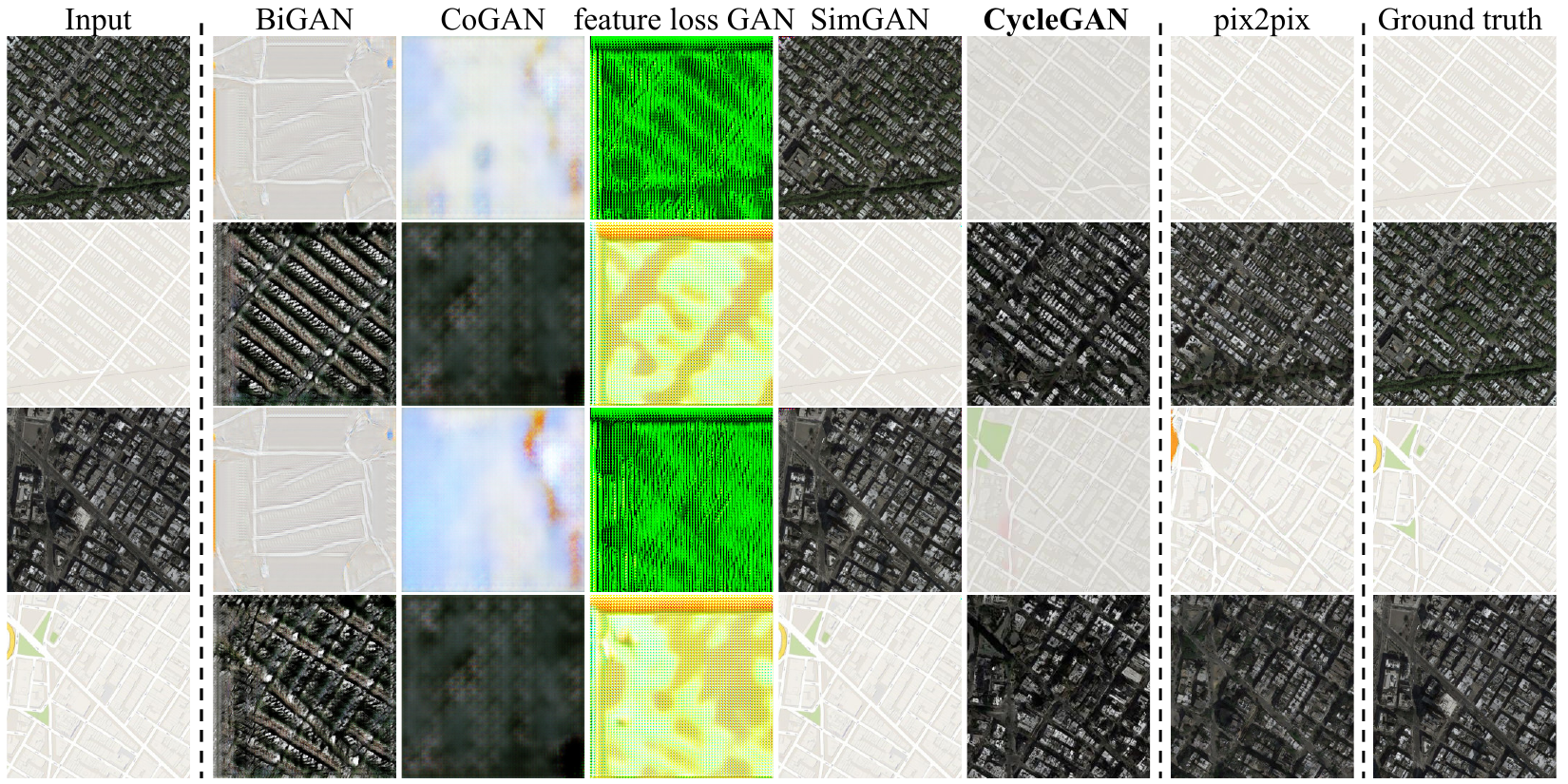} 
 \vspace{-4 mm}
\caption{Different methods for mapping aerial photos$\leftrightarrow$maps on Google Maps. From left to right: input, BiGAN/ALI~\cite{donahue2016adversarial,dumoulin2016adversarially}, CoGAN~\cite{liu2016coupled},  feature loss + GAN, SimGAN~\cite{shrivastava2016learning}, CycleGAN (ours), pix2pix~\cite{isola2016image} trained on paired data, and ground truth.}
\lblfig{maps_baseline}
\end{center}
 \vspace{-8 mm}

\end{figure*}

Using the same evaluation datasets and metrics as ``pix2pix''~\shortcite{isola2016image}, we compare our method against several baselines both qualitatively and quantitatively. The tasks include semantic labels$\leftrightarrow$photo on the Cityscapes dataset~\cite{Cordts2016Cityscapes}, and map$\leftrightarrow$aerial photo on data scraped from Google Maps. We also perform ablation study on the full loss function.
	
\vspace{-1 mm}
\subsubsection{Evaluation Metrics}
\vspace{-1 mm}

\hspace{4mm} {\bf AMT perceptual studies} On the map$\leftrightarrow$aerial photo task, we run ``real vs fake" perceptual studies on Amazon Mechanical Turk (AMT) to assess the realism of our outputs. We follow the same perceptual study protocol from Isola et al.~\shortcite{isola2016image}, except we only gather data from $25$ participants per algorithm we tested. Participants were shown a sequence of pairs of images, one a real photo or map and one fake (generated by our algorithm or a baseline), and asked to click on the image they thought was real. The first $10$ trials of each session were practice and feedback was given as to whether the participant's response was correct or incorrect. The remaining $40$ trials were used to assess the rate at which each algorithm fooled participants. Each session only tested a single algorithm, and participants were only allowed to complete a single session. The numbers we report here are not directly comparable to those in~\cite{isola2016image} as our ground truth images were processed slightly differently
\footnote{We train all the models on $256\times256$ images while in pix2pix~\cite{isola2016image}, the model was trained on $256\times256$ patches of $512\times512$ images, and run convolutionally on the $512\times512$ images at test time. We choose $256\times256$ in our experiments as many baselines cannot scale up to high-resolution images, and CoGAN cannot be tested fully convolutionally.} and the participant pool we tested may be differently distributed from those tested in~\cite{isola2016image} (due to running the experiment at a different date and time). Therefore, our numbers should only be used to compare our current method against the baselines (which were run under identical conditions), rather than against~\cite{isola2016image}. 

{\bf FCN score} Although perceptual studies may be the gold standard for assessing graphical realism, we also seek an automatic quantitative measure that does not require human experiments. For this, we adopt the ``FCN score" from~\cite{isola2016image}, and use it to evaluate the Cityscapes labels$\rightarrow$photo task. The FCN metric evaluates how interpretable the generated photos are according to an off-the-shelf semantic segmentation algorithm (the fully-convolutional network, FCN, from~\cite{long2015fully}). The FCN predicts a label map for a generated photo. This label map can then be compared against the input ground truth labels using standard semantic segmentation metrics described below. The intuition is that if we generate a photo from a label map of ``car on the road", then we have succeeded if the FCN applied to the generated photo detects ``car on the road".

{\bf Semantic segmentation metrics} To evaluate the performance of photo$\rightarrow$labels, we use the standard metrics from the Cityscapes benchmark~\cite{Cordts2016Cityscapes}, including per-pixel accuracy, per-class accuracy, and mean class Intersection-Over-Union (Class IOU)~\cite{Cordts2016Cityscapes}.

\vspace{-1 mm}
\subsubsection{Baselines}
\vspace{-1 mm}

\hspace{4mm} {\bf CoGAN~\cite{liu2016coupled}} This method learns one GAN generator for domain $X$ and one for domain $Y$, with tied weights on the first few layers for shared latent representations. Translation from $X$ to $Y$ can be achieved by finding a latent representation that generates image $X$ and then rendering this latent representation into style $Y$.

{\bf SimGAN~\cite{shrivastava2016learning}} Like our method, Shrivastava et al.\shortcite{shrivastava2016learning} uses an adversarial loss to train a translation from $X$ to $Y$. The regularization term $\norm{x-G(x)}_1$ i s used to penalize making large changes at pixel level.

{\bf Feature loss + GAN} We also test a variant of SimGAN~\cite{shrivastava2016learning} where the L1 loss is computed over deep image features using a pretrained network (VGG-16 \texttt{relu4\_2}~\cite{simonyan2014very}), rather than over RGB pixel values. Computing distances in deep feature space, like this, is also sometimes referred to as using a ``perceptual loss"~\cite{dosovitskiy2016generating,johnson2016perceptual}.

{\bf BiGAN/ALI~\cite{dumoulin2016adversarially,donahue2016adversarial}} Unconditional GANs~\cite{goodfellow2014generative} learn a generator $G: Z\rightarrow X$, that maps a random noise $z$ to an image $x$. The BiGAN~\cite{dumoulin2016adversarially} and ALI~\cite{donahue2016adversarial} propose to also learn the inverse mapping function $F: X\rightarrow Z$. Though they were originally designed for mapping a latent vector $z$ to an image $x$, we implemented the same objective for mapping a source image $x$ to a target image $y$.

{\bf pix2pix~\cite{isola2016image}} We also compare against pix2pix~\cite{isola2016image}, which is trained on paired data, to see how close we can get to this ``upper bound" without using any paired data. 

For a fair comparison, we implement all the baselines using the same architecture and details as our method, except for CoGAN~\cite{liu2016coupled}. CoGAN builds on generators that produce images from a shared latent representation, which is incompatible with our image-to-image network. We use the public \href{https://github.com/mingyuliutw/CoGAN}{implementation} of CoGAN instead.

\begin{table}
\centering
\scalebox{0.75} {
\begin{tabular}{lcc}
 & \textbf{Map $\rightarrow$ Photo}  &  \textbf{Photo $\rightarrow$ Map} \\ 
\textbf{Loss} & \% Turkers labeled \emph{real} & \% Turkers labeled \emph{real} \\
\hline
CoGAN~\cite{liu2016coupled} & 0.6\% $\pm$ 0.5\% & 0.9\% $\pm$ 0.5\% \\
BiGAN/ALI~\cite{dumoulin2016adversarially,donahue2016adversarial} & 2.1\% $\pm$ 1.0\% & 1.9\% $\pm$ 0.9\% \\
SimGAN~\cite{shrivastava2016learning} & 0.7\% $\pm$ 0.5\% & 2.6\% $\pm$ 1.1\% \\
Feature loss + GAN & 1.2\% $\pm$ 0.6\% & 0.3\% $\pm$ 0.2\% \\
CycleGAN (ours)  & {\bf 26.8\% $\pm$ 2.8\%} &  {\bf 23.2\% $\pm$ 3.4\%} \\
\end{tabular} }
\vspace{-0.1in}
\caption {AMT ``real vs fake" test on maps$\leftrightarrow$aerial photos at $256\times 256$ resolution.}
\vspace{-0.1in}
\lbltbl{AMT_map2sat}
\end{table}

\begin{table}
\centering
\vspace{-2mm}
\scalebox{0.75} {
\begin{tabular}{lcccc}
 & & & \\
\textbf{Loss} & \textbf{Per-pixel acc.} & \textbf{Per-class acc.} & \textbf{Class IOU} \\ \hline
CoGAN~\cite{liu2016coupled} & 0.40 & 0.10 & 0.06 \\
BiGAN/ALI~\cite{dumoulin2016adversarially,donahue2016adversarial} & 0.19 & 0.06 & 0.02 \\
SimGAN~\cite{shrivastava2016learning} & 0.20 & 0.10 & 0.04 \\
Feature loss + GAN & 0.06 & 0.04 & 0.01 \\
CycleGAN (ours) & {\bf 0.52} & {\bf 0.17} & {\bf 0.11} \\ \hline
pix2pix~\cite{isola2016image} & 0.71 & 0.25 & 0.18 \\ 
\end{tabular} }
\vspace{-0.1in}
\caption {FCN-scores for different methods, evaluated on Cityscapes labels$\rightarrow$photo.}
\lbltbl{labels_to_image_results}
\vspace{-0.2in}
\end{table}

\begin{table}
\centering
\scalebox{0.75} {
\begin{tabular}{lccccc}
 & & & \\
\textbf{Loss} & \textbf{Per-pixel acc.} & \textbf{Per-class acc.} & \textbf{Class IOU} \\ \hline
CoGAN~\cite{liu2016coupled} & 0.45 & 0.11 & 0.08\\
BiGAN/ALI~\cite{dumoulin2016adversarially,donahue2016adversarial} & 0.41 & 0.13 & 0.07 \\
SimGAN~\cite{shrivastava2016learning} & 0.47 & 0.11 & 0.07\\
Feature loss + GAN & 0.50  & 0.10 & 0.06 \\
CycleGAN (ours) & {\bf 0.58} & {\bf 0.22} & {\bf 0.16}\\\hline
pix2pix~\cite{isola2016image} & 0.85 & 0.40 & 0.32 \\
\end{tabular} }
\vspace{-0.1in}
\caption {Classification performance of photo$\rightarrow$labels for different methods on cityscapes.}
\vspace{-0.1in}
\lbltbl{image_to_labels_results}
\end{table}

\begin{figure*}[ht]
\begin{center}
\includegraphics[width=1.0\linewidth]{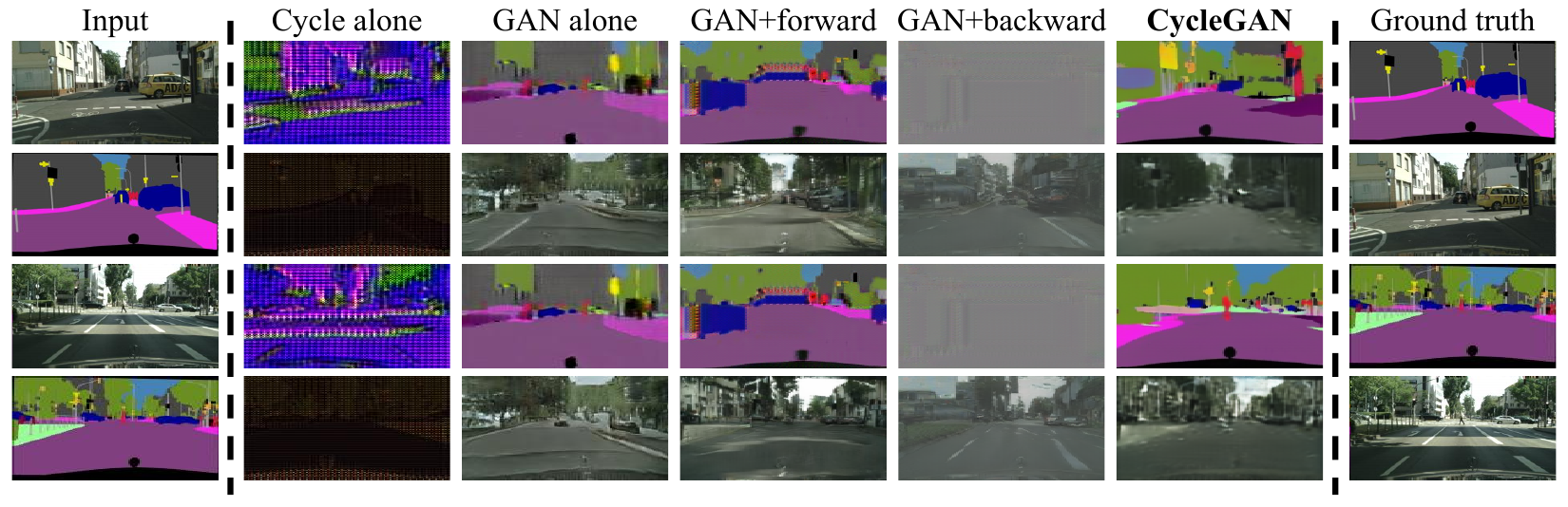}
\end{center}
 \vspace{-5 mm}
 \caption{Different variants of our method for mapping labels$\leftrightarrow$photos trained on cityscapes. From left to right: input, cycle-consistency loss alone, adversarial loss alone, GAN + forward cycle-consistency loss ($F(G(x)) \approx x$), GAN + backward cycle-consistency loss ($G(F(y)) \approx y$), CycleGAN (our full method), and ground truth. Both  \emph{Cycle alone} and \emph{GAN + backward} fail to produce images similar to the target domain. \emph{GAN alone} and \emph{GAN + forward} suffer from mode collapse, producing identical label maps regardless of the input photo.} 
\lblfig{ablation} 
 \vspace{-5 mm}
\end{figure*}

\begin{table}
\centering
\scalebox{0.75} {
\begin{tabular}{lcccc}
 & & & \\
\textbf{Loss} & \textbf{Per-pixel acc.} & \textbf{Per-class acc.} & \textbf{Class IOU} \\ \hline
Cycle alone & 0.22 & 0.07 & 0.02 \\
GAN alone & 0.51 & 0.11 & 0.08 \\
GAN + forward cycle & {\bf 0.55} & {\bf 0.18} & {\bf 0.12} \\ 
GAN + backward cycle & 0.39 & 0.14 & 0.06 \\ 
CycleGAN (ours) & 0.52 & 0.17 & 0.11 \\  
\end{tabular} }
\vspace{-0.1in}
\caption {Ablation study: FCN-scores for different variants of our method, evaluated on Cityscapes labels$\rightarrow$photo.}
\vspace{-0.2in}
\lbltbl{loss_fcn}
\end{table}

\begin{table}
\centering
\scalebox{0.75} {
\begin{tabular}{lcccc}
 & & & \\
\textbf{Loss} & \textbf{Per-pixel acc.} & \textbf{Per-class acc.} & \textbf{Class IOU} \\ \hline
Cycle alone & 0.10 & 0.05 & 0.02 \\
GAN alone & 0.53 & 0.11 & 0.07 \\
GAN + forward cycle & 0.49 & 0.11 & 0.07 \\ 
GAN + backward cycle & 0.01 & 0.06 & 0.01 \\ 
CycleGAN (ours)& {\bf 0.58} & {\bf 0.22} & {\bf 0.16} \\  
\end{tabular} }
\vspace{-0.1in}
\caption {Ablation study: classification performance of photo$\rightarrow$labels for different losses, evaluated on Cityscapes. }
\vspace{-0.2in}
\lbltbl{loss_clf}
\end{table}

\subsubsection{Comparison against baselines}
As can be seen in \reffig{cityscapes_baseline} and \reffig{maps_baseline}, we were unable to achieve compelling results with any of the baselines. Our method, on the other hand, can produce translations that are often of similar quality to the fully supervised pix2pix.

\reftbl{AMT_map2sat} reports performance regarding the AMT perceptual realism task. Here, we see that our method can fool participants on around a quarter of trials, in both the maps$\rightarrow$aerial photos direction and the aerial photos$\rightarrow$maps direction at $256\times 256$ resolution\footnote{We also train CycleGAN and pix2pix at $512\times 512$ resolution, and observe the comparable performance:  maps$\rightarrow$aerial photos: CycleGAN: $37.5\% \pm 3.6\%$ and pix2pix: $33.9\% \pm 3.1\%$; aerial photos$\rightarrow$maps: CycleGAN: $16.5\% \pm 4.1\%$ and pix2pix: $8.5\% \pm 2.6\%$}. All the baselines almost never fooled participants.

\reftbl{labels_to_image_results} assesses the performance of the labels$\rightarrow$photo task on the Cityscapes and \reftbl{image_to_labels_results} evaluates the opposite mapping (photos$\rightarrow$labels). In both cases, our method again outperforms the baselines.

\vspace{-1 mm}
\subsubsection{Analysis of the loss function}
\lblsec{ablation}
\vspace{-1 mm}

In \reftbl{loss_fcn} and \reftbl{loss_clf}, we compare against ablations of our full loss. Removing the GAN loss substantially degrades results, as does removing the cycle-consistency loss. We therefore conclude that both terms are critical to our results. We also evaluate our method with the cycle loss in only one direction: GAN + forward cycle loss $\mathbb{E}_{x\sim p_{\text{data}}(x)}[\norm{F(G(x))-x}_1]$, or GAN + backward cycle loss $\mathbb{E}_{y\sim p_{\text{data}}(y)}[\norm{G(F(y))-y}_1]$ (\refeq{cycle}) and find that it often incurs training instability and causes mode collapse, especially for the direction of the mapping that was removed. \reffig{ablation} shows several qualitative examples. 

\vspace{-1 mm}
\subsubsection{Image reconstruction quality}
\vspace{-1 mm}

In \reffig{reconstruction}, we show a few random samples of the reconstructed images $F(G(x))$. We observed that the reconstructed images were often close to the original inputs $x$, at both training and testing time, even in cases where one domain represents significantly more diverse information, such as map$\leftrightarrow$aerial photos.

\begin{figure}[t]
\begin{center}
\includegraphics[width=1.0\linewidth]{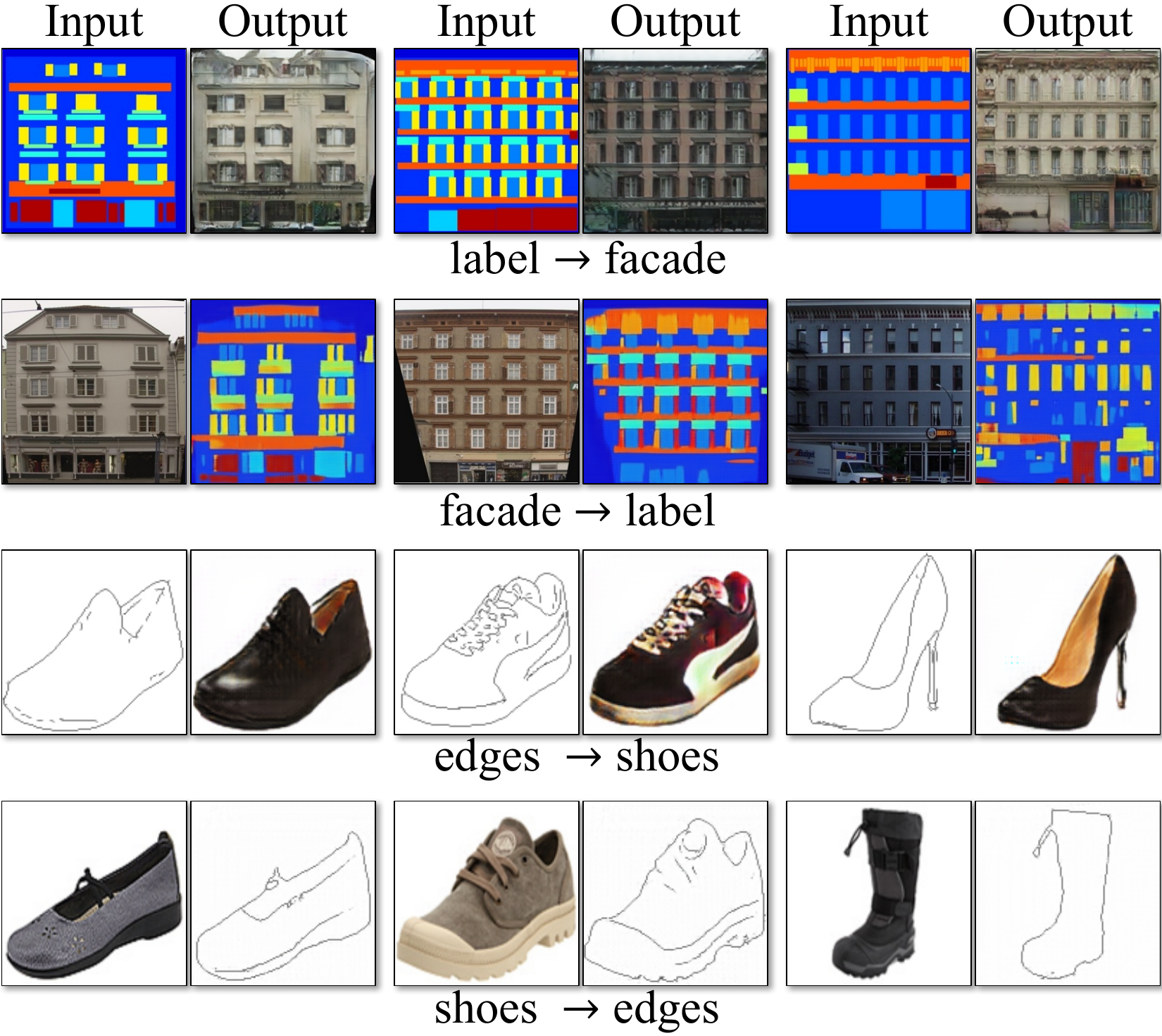}
\end{center}
 \vspace{-5 mm}
 \caption{Example results of CycleGAN on paired datasets used in ``pix2pix''~\cite{isola2016image} such as architectural labels$\leftrightarrow$photos and edges$\leftrightarrow$shoes.}
\lblfig{facades_and_sketches}
 \vspace{-5 mm}
\end{figure}

\vspace{-1 mm}
\subsubsection{Additional results on paired datasets}
\vspace{-1 mm}
\reffig{facades_and_sketches} shows some example results on other paired datasets used in ``pix2pix''~\cite{isola2016image}, such as architectural labels$\leftrightarrow$photos from the CMP Facade Database~\cite{Tylecek13}, and edges$\leftrightarrow$shoes from the UT Zappos50K
dataset~\cite{yu2014fine}. The image quality of our results is close to those produced by the fully supervised pix2pix while our method learns the mapping without paired supervision.

\vspace{-1 mm}
\subsection{Applications}
\lblsec{application}
\vspace{-1 mm}

We demonstrate our method on several applications where paired training data does not exist. Please refer to the appendix (\refsec{appendix}) for more details about the datasets. We observe that translations on training data are often more appealing than those on test data, and full results of all applications on both training and test data can be viewed on our project \href{https://junyanz.github.io/CycleGAN/}{website}.

{\bf Collection style transfer (\reffig{painting} and ~\reffig{painting2})}  We train the model on landscape photographs downloaded from Flickr and WikiArt. Unlike recent work on ``neural style transfer"~\cite{gatys2015neural}, our method learns to mimic the style of an entire \emph{collection} of artworks, rather than transferring the style of a single selected piece of art. Therefore, we can learn to generate photos in the style of, e.g., Van Gogh, rather than just in the style of Starry Night. The size of the dataset for each artist/style was $526$, $1073$, $400$, and $563$ for Cezanne, Monet, Van Gogh, and Ukiyo-e.

{\bf Object transfiguration (\reffig{big})} The model is trained to translate one object class from ImageNet~\cite{deng2009imagenet} to another (each class contains around $1000$ training images). Turmukhambetov et al.~\shortcite{turmukhambetov2015modeling} propose a subspace model to translate one object into another object of the same category, while our method focuses on object transfiguration between two visually similar categories. 

{\bf Season transfer (\reffig{big})}  The model is trained on $854$ winter photos and $1273$ summer photos of Yosemite downloaded from Flickr.

\begin{figure}[t]
\begin{center}
\includegraphics[width=1.0\linewidth]{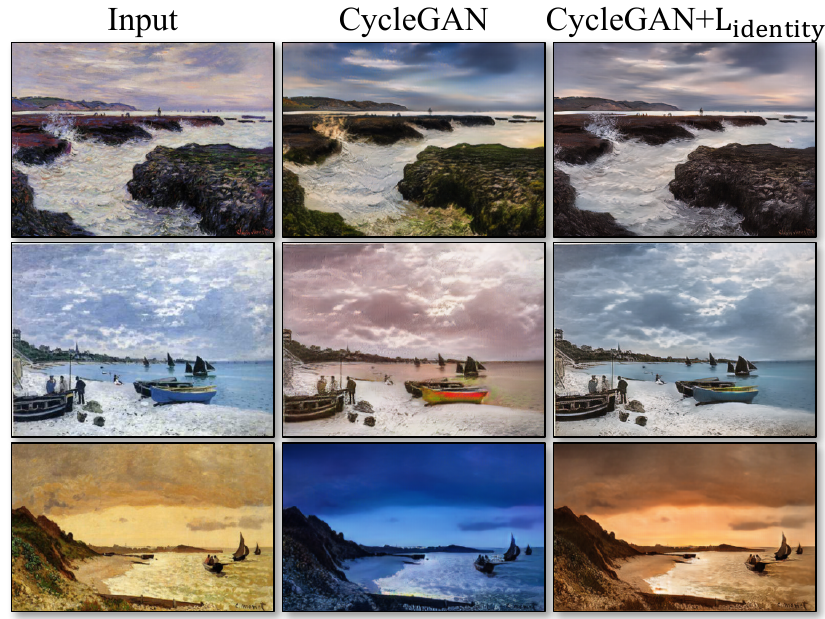}
\end{center}
 \vspace{-5 mm}
 \caption{The effect of the \textit{identity mapping loss} on Monet's painting$\rightarrow$ photos. From left to right: input paintings, CycleGAN without identity mapping loss, CycleGAN with identity mapping loss. The identity mapping loss helps preserve the color of the input paintings. }
\lblfig{identity}
 \vspace{-6mm}
\end{figure}

{\bf Photo generation from paintings (\reffig{painting2photo})} 
For painting$\rightarrow$photo, we find that it is helpful to introduce an additional loss to encourage the mapping to preserve color composition between the input and output. In particular, we adopt the technique of Taigman et al.~\cite{taigman2016unsupervised} and regularize the generator to be near an identity mapping when real samples of the target domain are provided as the input to the generator: i.e., $ \mathcal{L}_{\text{identity}}(G, F) = \mathbb{E}_{y\sim p_{\text{data}}(y)}[\norm{G(y) - y}_1]+\mathbb{E}_{x\sim p_{\text{data}}(x)}
    [\norm{F(x) - x}_1].$ 

Without $\mathcal{L}_{\text{identity}}$, the generator $G$ and $F$ are free to change the tint of input images when there is no need to. For example, when learning the mapping between Monet's paintings and Flickr photographs, the generator often maps paintings of daytime to photographs taken during sunset, because such a mapping may be equally valid under the adversarial loss and cycle consistency loss. 
The effect of this \emph{identity mapping loss} are shown in \reffig{identity}.

In \reffig{painting2photo}, we show additional results translating Monet's paintings to photographs. This figure and \reffig{identity} show results on paintings that were included in the \emph{training set}, whereas for all other experiments in the paper, we only evaluate and show test set results. Because the training set does not include paired data, coming up with a plausible translation for a training set painting is a nontrivial task. Indeed, since Monet is no longer able to create new paintings, generalization to unseen, ``test set", paintings is not a pressing problem.

{\bf Photo enhancement (\reffig{flower})} We show that our method can be used to generate photos with shallower depth of field. We train the model on flower photos downloaded from Flickr. The source domain consists of flower photos taken by smartphones, which usually have deep DoF due to a small aperture. The target contains photos captured by DSLRs with a larger aperture. Our model successfully generates photos with shallower depth of field from the photos taken by smartphones.

{\bf Comparison with Gatys et al.~\cite{gatys2015neural}}
In ~\reffig{gatys_style}, we compare our results with neural style transfer~\cite{gatys2015neural} on photo stylization. For each row, we first use two representative artworks as the style images for ~\cite{gatys2015neural}. Our method, on the other hand, can produce photos in the style of entire \emph{collection}. To compare against neural style transfer of an entire collection, we compute the average Gram Matrix across the target domain and use this matrix to transfer the ``average style" with Gatys et al~\cite{gatys2015neural}. 

\reffig{gatys_others} demonstrates similar comparisons for other translation tasks. 
We observe that Gatys et al.~\cite{gatys2015neural} requires finding target style images that closely match the desired output, but still often fails to produce photorealistic results, while our method succeeds to generate natural-looking results, similar to the target domain. 
\section{Limitations and Discussion}
Although our method can achieve compelling results in many cases, the results are far from uniformly positive. \reffig{failure} shows several typical failure cases. On translation tasks that involve color and texture changes, as many of those reported above, the method often succeeds. We have also explored tasks that require geometric changes, with little success. For example, on the task of dog$\rightarrow$cat transfiguration, the learned translation degenerates into making minimal changes to the input (\reffig{failure}). This failure might be caused by our generator architectures which are tailored for good performance on the appearance changes. Handling more varied and extreme transformations, especially geometric changes, is an important problem for future work.

Some failure cases are caused by the distribution characteristics of the training datasets. For example, our method  has got confused in the horse $\rightarrow$ zebra example (\reffig{failure}, right), because our model was trained on the {\it wild horse} and {\it zebra} synsets of ImageNet, which does not contain images of a person riding a horse or zebra.

We also observe a lingering gap between the results achievable with paired training data and those achieved by our unpaired method. In some cases, this gap may be very hard -- or even impossible -- to close: for example, our method sometimes permutes the labels for tree and building in the output of the photos$\rightarrow$labels task. Resolving this ambiguity may require some form of weak semantic supervision. Integrating weak or semi-supervised data may lead to substantially more powerful translators, still at a fraction of the annotation cost of the fully-supervised systems.

Nonetheless, in many cases completely unpaired data is plentifully available and should be made use of. This paper pushes the boundaries of what is possible in this ``unsupervised" setting.

{\bf Acknowledgments:} We thank Aaron Hertzmann, Shiry Ginosar, Deepak Pathak, Bryan Russell, Eli Shechtman, Richard Zhang, and Tinghui Zhou for many helpful comments. This work was supported in part by NSF SMA-1514512, NSF IIS-1633310, a Google Research Award, Intel Corp, and hardware donations from NVIDIA. JYZ is supported by the Facebook Graduate Fellowship and TP is supported by the Samsung Scholarship. The photographs used for style transfer were taken by AE, mostly in France.

\begin{figure*}[h]
\begin{center}
\includegraphics[width=1.0\linewidth]{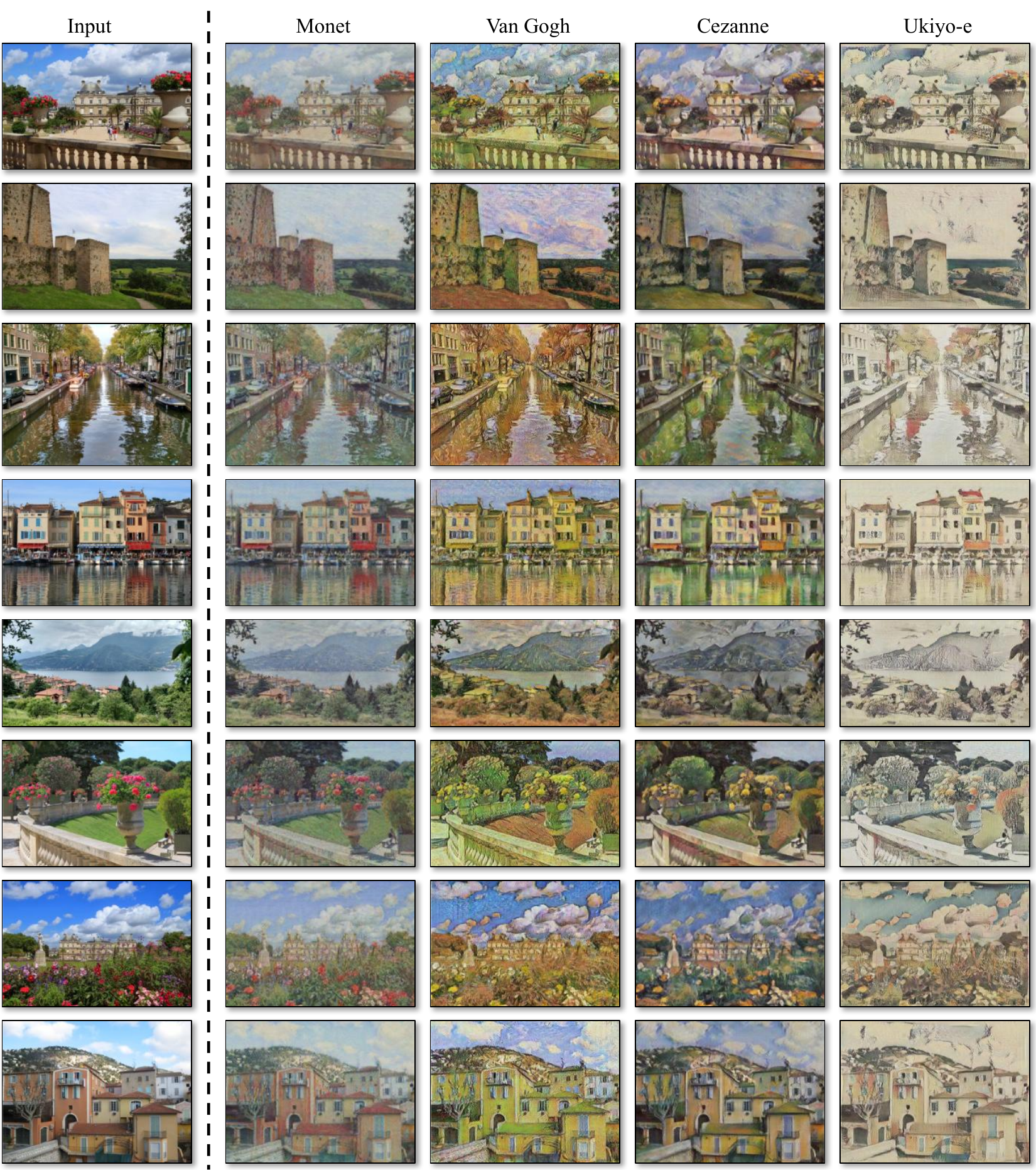}
\end{center}
 \caption{Collection style transfer I: we transfer input images into the artistic styles of Monet, Van Gogh, Cezanne, and Ukiyo-e. Please see our \href{https://junyanz.github.io/CycleGAN/}{website} for additional examples.}
\lblfig{painting}
\end{figure*}

\begin{figure*}[h]
\begin{center}
\includegraphics[width=1.0\linewidth]{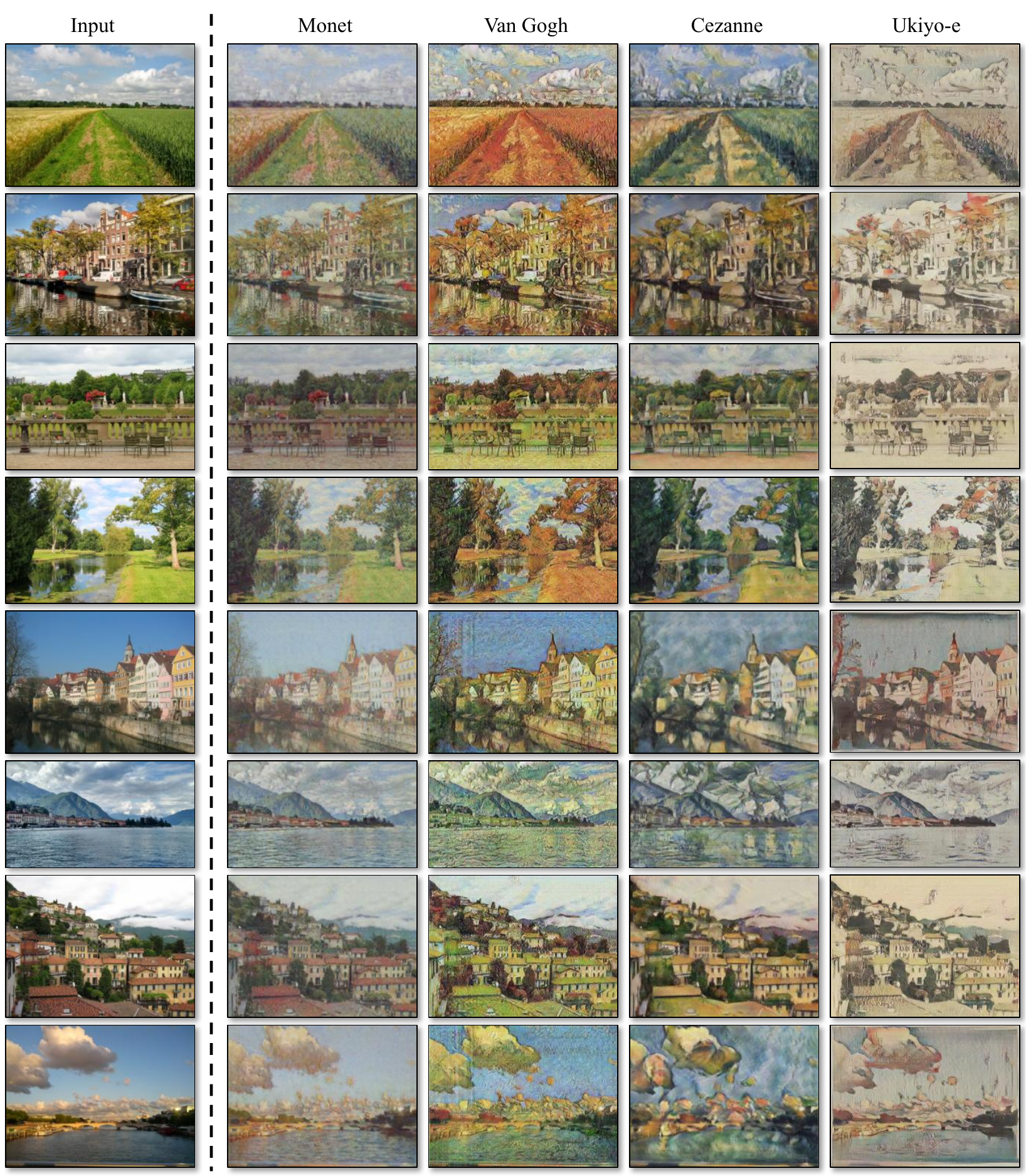}
\end{center}
 \caption{Collection style transfer II: we transfer input images into the artistic styles of Monet, Van Gogh, Cezanne, Ukiyo-e. Please see our \href{https://junyanz.github.io/CycleGAN/}{website} for additional examples.}
\lblfig{painting2}
\end{figure*}

\begin{figure*}[h]
\begin{center}
\includegraphics[width=1\linewidth]{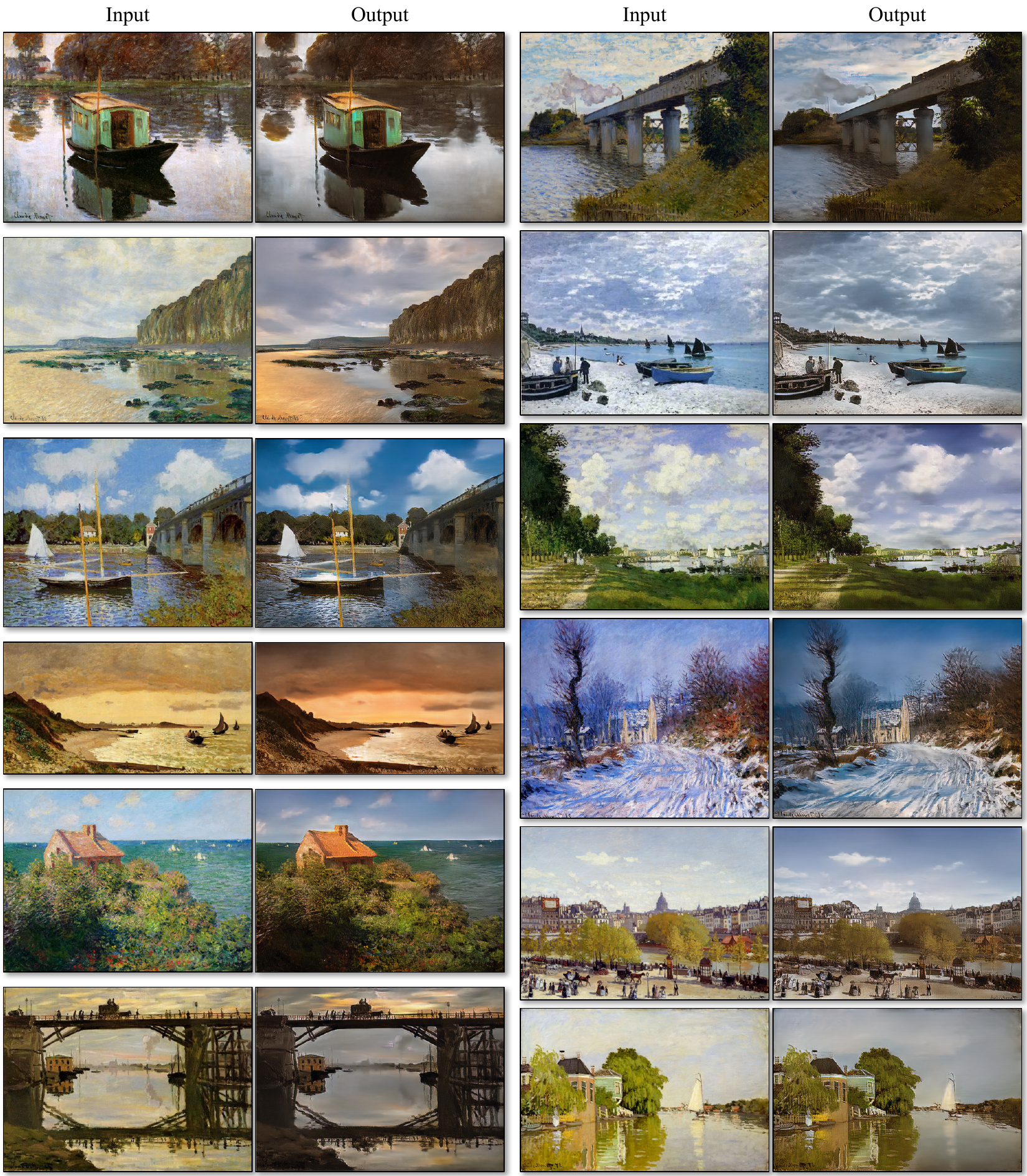}
\end{center}
 \caption{Relatively successful results on mapping Monet's paintings to a photographic style. Please see our \href{https://junyanz.github.io/CycleGAN/}{website} for additional examples.}
\lblfig{painting2photo}
 \vspace{24mm}
\end{figure*}

\begin{figure*}[t]
\begin{center}
\includegraphics[width=1.0\linewidth]{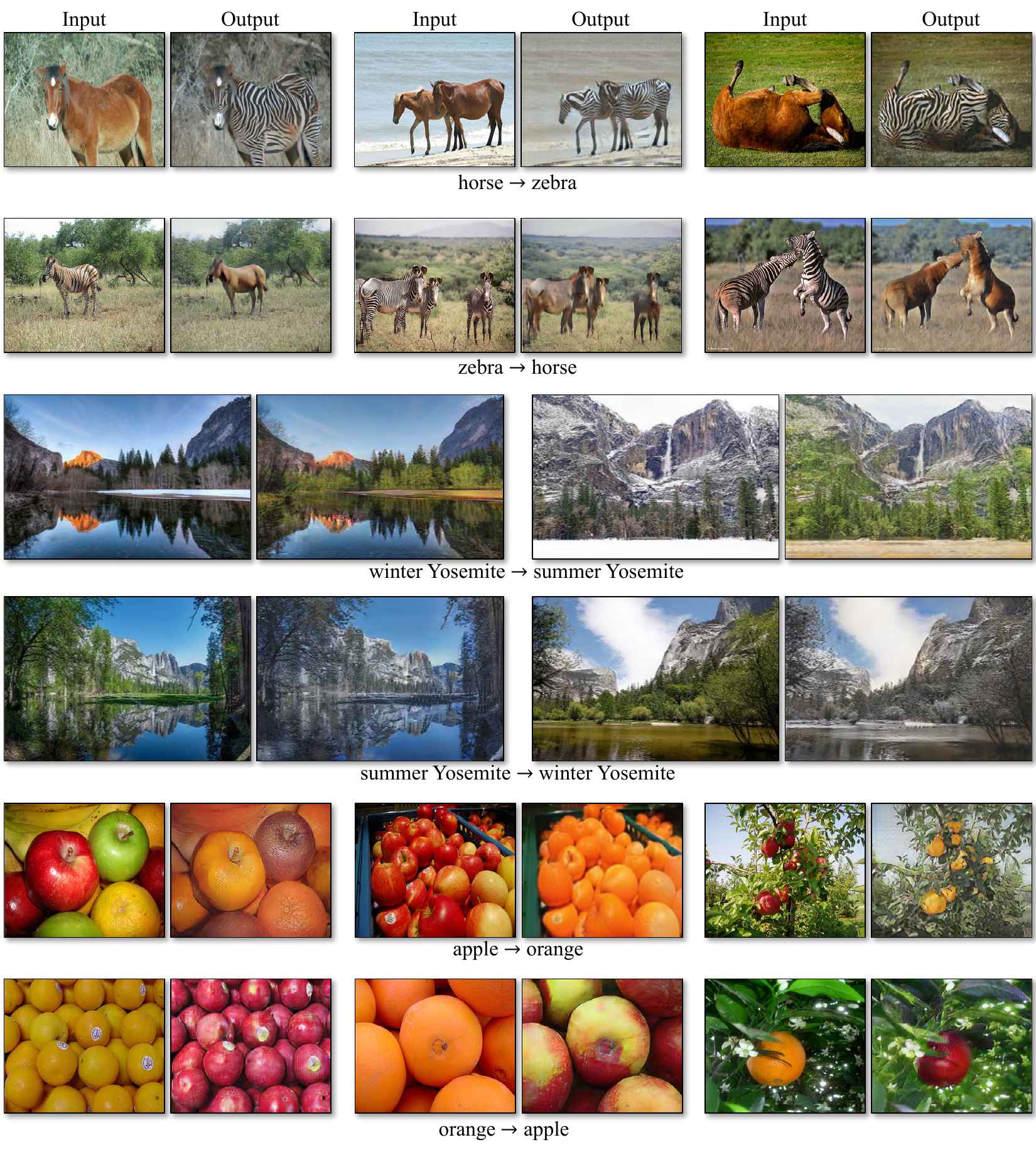}
\end{center}
 \vspace{-3 mm}
  \caption{Our method applied to several translation problems. These images are selected as relatively successful results -- please see our \href{https://junyanz.github.io/CycleGAN/}{website} for more comprehensive and random results. In the top two rows, we show results on object transfiguration between horses and zebras, trained on 939 images from the \emph{wild horse} class and 1177 images from the \emph{zebra} class in Imagenet~\cite{deng2009imagenet}. Also check out the horse$\rightarrow$zebra demo \href{https://youtu.be/9reHvktowLY}{video}. The middle two rows show results on season transfer, trained on winter and summer photos of Yosemite from Flickr. In the bottom two rows, we train our method on 996 \emph{apple} images and 1020 \emph{navel orange} images from ImageNet.}
\lblfig{big}
\end{figure*}

\begin{figure*}[t]
\begin{center}
\includegraphics[width=1.0\linewidth]{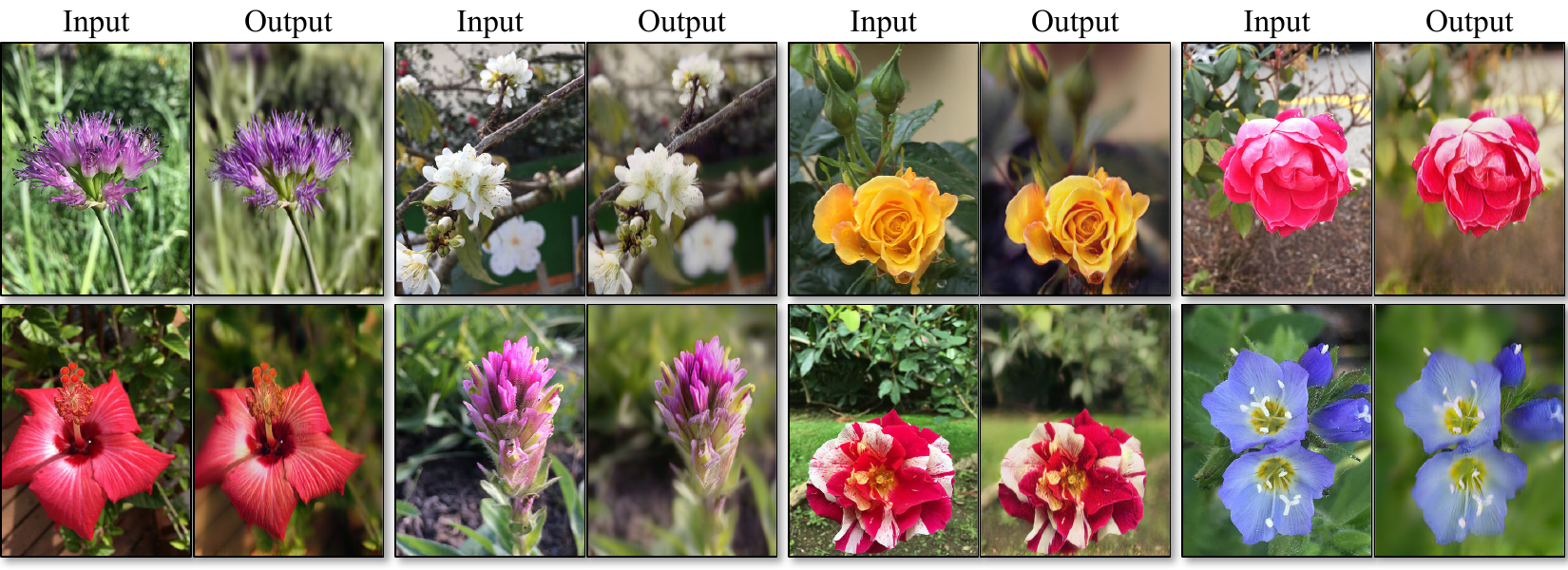}
\end{center}
 \vspace{-4 mm}
 \caption{Photo enhancement: mapping from a set of smartphone snaps to professional DSLR photographs, the system often learns to produce shallow focus. Here we show some of the most successful results in our test set -- average performance is considerably worse. Please see our \href{https://junyanz.github.io/CycleGAN/}{website} for more comprehensive and random examples. }
\lblfig{flower}
 \vspace{-4 mm}
\end{figure*}

\begin{figure*}[t]
\begin{center}
\includegraphics[width=1.0\linewidth]{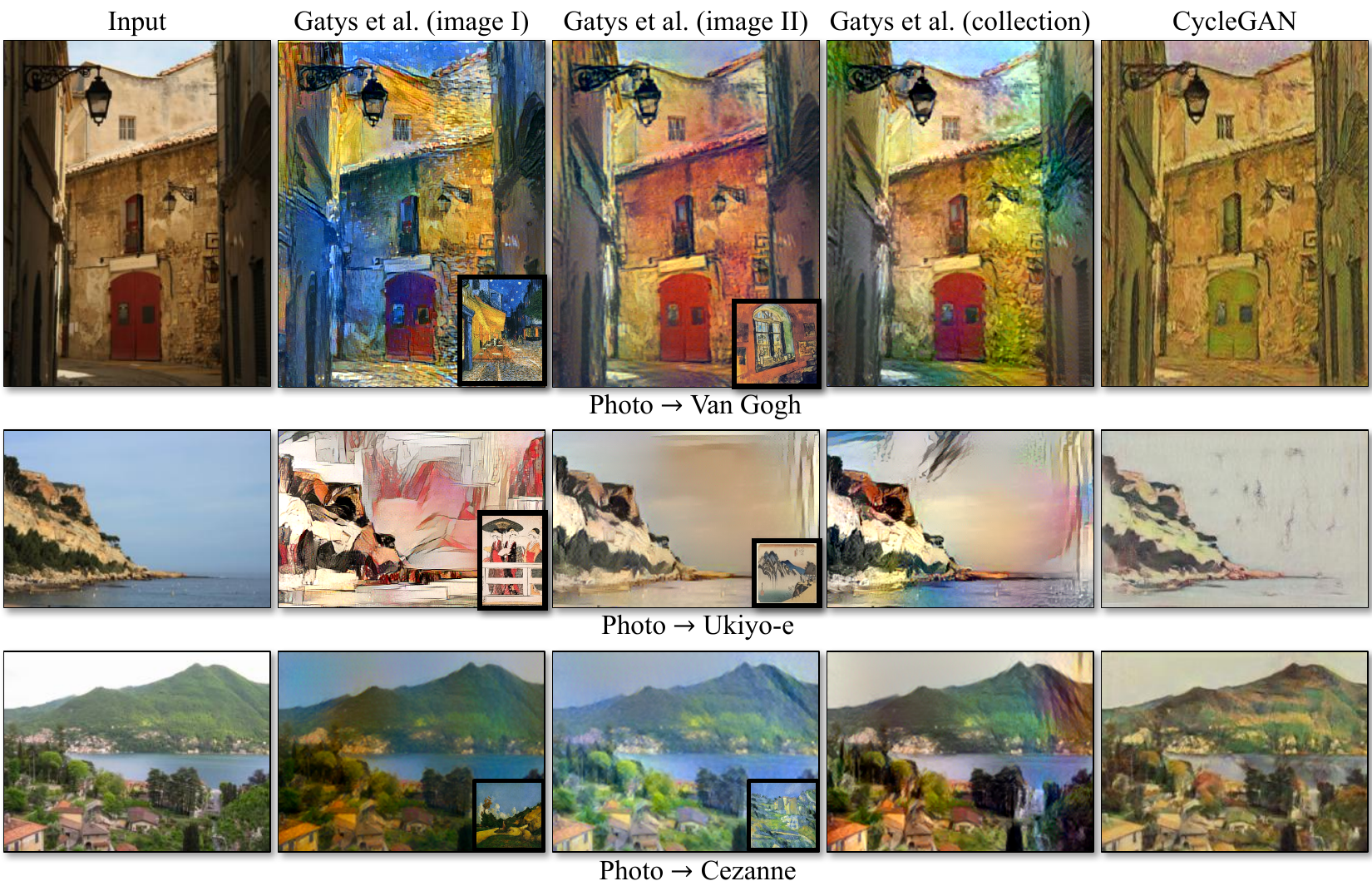}
\end{center}
 \vspace{-4 mm}
 \caption{We compare our method with neural style transfer~\cite{gatys2015neural} on photo stylization. Left to right: input image, results from Gatys et al.~\cite{gatys2015neural} using two different representative artworks as style images, results from Gatys et al.~\cite{gatys2015neural} using the entire collection of the artist, and CycleGAN (ours).}
\lblfig{gatys_style}
 \vspace{-4 mm}
\end{figure*}

\begin{figure*}[t]
\begin{center}
\includegraphics[width=1.0\linewidth]{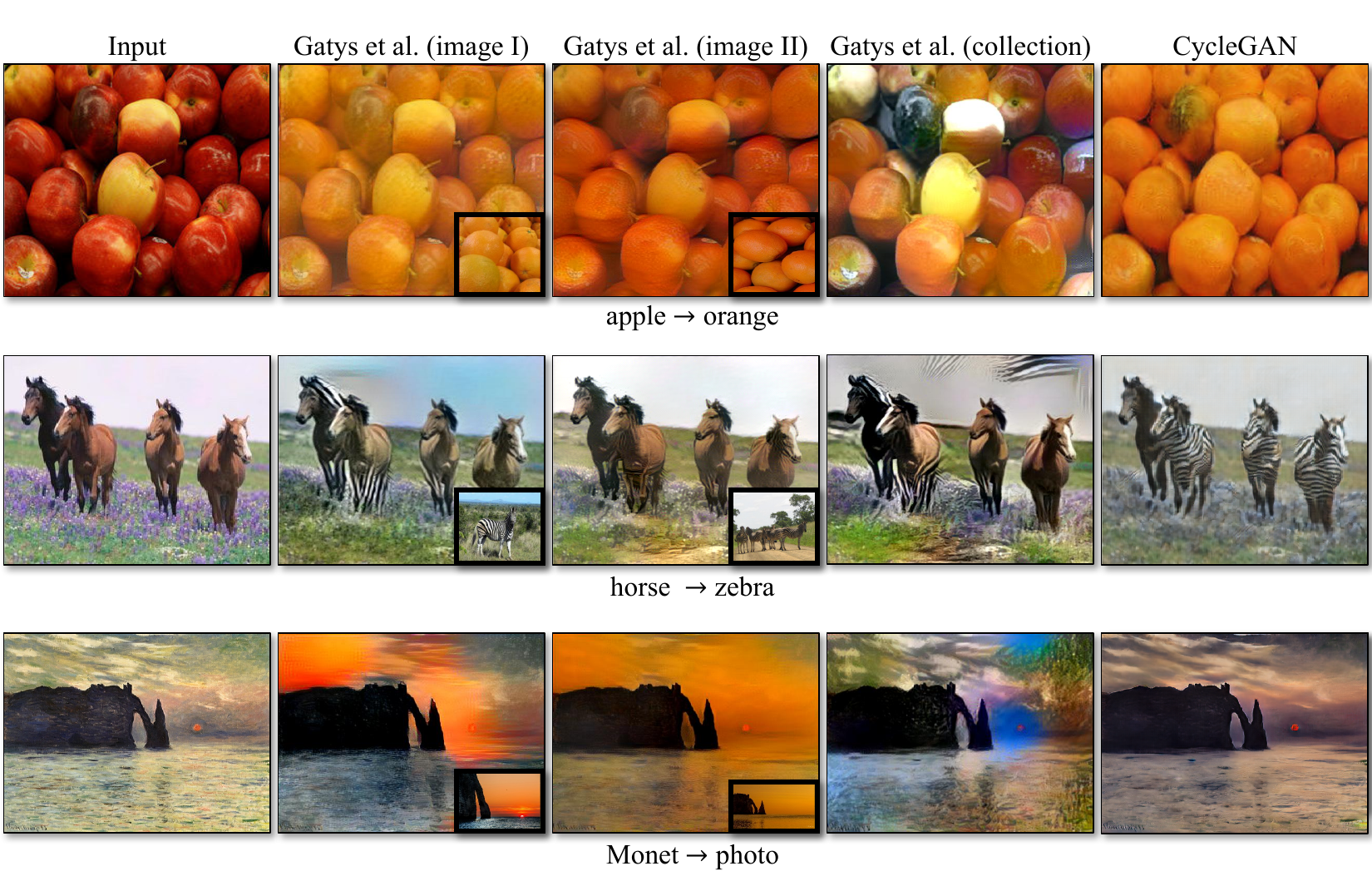}
\end{center}
 \vspace{-4 mm}
 \caption{We compare our method with neural style transfer~\cite{gatys2015neural} on various applications. From top to bottom:  apple$\rightarrow$orange, horse$\rightarrow$zebra, and Monet$\rightarrow$photo. Left to right: input image, results from Gatys et al.~\cite{gatys2015neural} using two different images as style images, results from Gatys et al.~\cite{gatys2015neural} using all the images from the target domain, and CycleGAN (ours).} 
\lblfig{gatys_others}
 \vspace{-4 mm}
\end{figure*}

\begin{figure*}[t]
\begin{center}
  \centering
\subfloat{\includegraphics[width =1.0\linewidth]{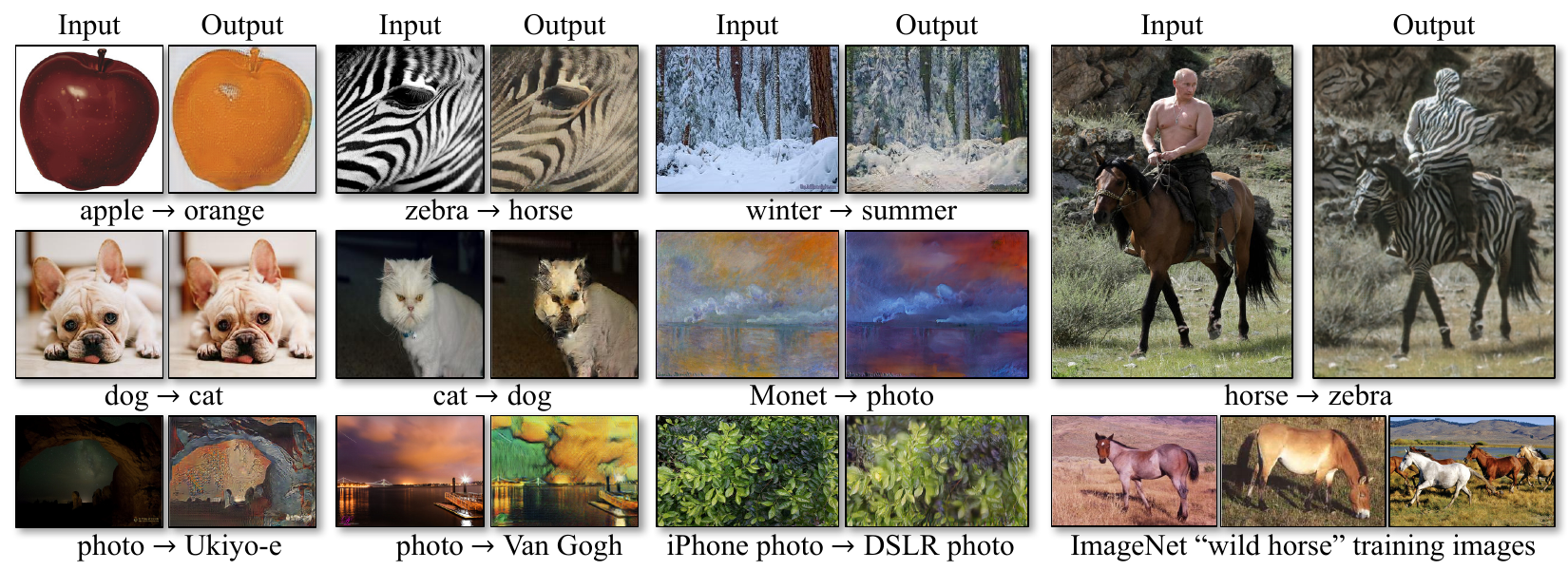}} 

\end{center}
 \vspace{-4 mm}
 \caption{Typical failure cases of our method. Left: in the task of dog$\rightarrow$cat transfiguration, CycleGAN can only make minimal changes to the input.
 Right: CycleGAN also fails in this horse $\rightarrow$ zebra example as our model has not seen images of horseback riding during training.
 Please see our \href{https://junyanz.github.io/CycleGAN/}{website} for more  comprehensive results.}
  \vspace{-2 mm}
\lblfig{failure}
\end{figure*}

\clearpage

\bibliographystyle{ieee}
\bibliography{main}
\clearpage 
\section{Appendix}
\lblsec{appendix}

\subsection{Training details}
We train our networks from scratch,  with a learning rate of $0.0002$. In practice, we divide the objective by $2$ while optimizing $D$, which slows down the rate at which $D$ learns, relative to the rate of $G$. We keep the same learning rate for the first $100$ epochs and linearly decay the rate to zero over the next $100$ epochs. Weights are initialized from a Gaussian distribution $\mathcal{N}(0, 0.02)$.

{\bf Cityscapes label$\leftrightarrow$Photo} $2975$ training images from the Cityscapes training set \cite{Cordts2016Cityscapes} with image size $128 \times 128$. We used the Cityscapes val set for testing.

{\bf Maps$\leftrightarrow$aerial photograph} $1096$ training images were scraped from Google Maps~\cite{isola2016image} with image size $256 \times 256$. Images were sampled from in and around New York City. Data was then split into train and test about the median latitude of the sampling region (with a buffer region added to ensure that no training pixel appeared in the test set).

{\bf Architectural facades labels$\leftrightarrow$photo} $400$ training images from the CMP Facade Database~\cite{Tylecek13}.

{\bf Edges$\rightarrow$shoes} around $50,000$ training images from UT Zappos50K dataset~\cite{yu2014fine}. The model was trained for $5$ epochs. 

{\bf Horse$\leftrightarrow$Zebra and Apple$\leftrightarrow$Orange} We downloaded the images from ImageNet~\cite{deng2009imagenet} using keywords \textit{wild horse}, \textit{zebra}, \textit{apple}, and \textit{navel orange}. The images were scaled to $256 \times 256$ pixels. The training set size of each class: $939$ (horse), $1177$ (zebra), $996$ (apple), and $1020$ (orange).

{\bf Summer$\leftrightarrow$Winter Yosemite} The images were downloaded using Flickr API with the tag \textit{yosemite} and the \textit{datetaken} field. Black-and-white photos were pruned. The images were scaled to $256 \times 256$ pixels. The training size of each class: $1273$ (summer) and $854$ ( winter).

{\bf Photo$\leftrightarrow$Art for style transfer} The art images were downloaded from Wikiart.org. Some artworks that were sketches or too obscene were pruned by hand. The photos were downloaded from Flickr using the combination of tags \textit{landscape} and \textit{landscapephotography}. Black-and-white photos were pruned. The images were scaled to $256 \times 256$ pixels. The training set size of each class was $1074$ (Monet), $584$ (Cezanne), $401$ (Van Gogh),  $1433$ (Ukiyo-e), and $6853$ (Photographs). The Monet dataset was particularly pruned to include only landscape paintings, and the Van Gogh dataset  included only his later works that represent his most recognizable artistic style. 

{\bf Monet's paintings$\rightarrow$photos} To achieve high resolution while conserving memory, we used random square crops of the original images for training. To generate results, we passed images of width $512$ pixels with correct aspect ratio to the generator network as input. The weight for the identity mapping loss was $0.5\lambda$ where $\lambda$ was the weight for cycle consistency loss. We set $\lambda=10$.

{\bf Flower photo enhancement} Flower images taken on smartphones were downloaded from Flickr by searching for the photos taken by 
\textit{Apple iPhone 5, 5s, or 6}, with search text \textit{flower}. DSLR images with shallow DoF were also downloaded from Flickr by search tag \textit{flower, dof}. The images were scaled to $360$ pixels by width. The identity mapping loss of weight $0.5\lambda$ was used. The training set size of the smartphone and DSLR dataset were $1813$ and $3326$, respectively. We set $\lambda=10$.

\subsection{Network architectures}
We provide both \href{https://github.com/junyanz/pytorch-CycleGAN-and-pix2pix}{PyTorch} and \href{https://github.com/junyanz/CycleGAN}{Torch} implementations. 

{\bf Generator architectures}
We adopt our architectures from Johnson et al.~\shortcite{johnson2016perceptual}. We use $6$ residual blocks for $128\times128$ training images, and $9$ residual blocks for $256\times 256$ or higher-resolution training images. Below, we follow the naming convention used in the Johnson et al.'s Github \href{https://github.com/jcjohnson/fast-neural-style}{repository}.

Let \texttt{c7s1-k} denote a $7\times7$ Convolution-InstanceNorm-ReLU layer with $k$ filters and stride $1$. \texttt{dk} denotes a $3\times3$ Convolution-InstanceNorm-ReLU layer with $k$ filters and stride $2$. Reflection padding was used to reduce artifacts. \texttt{Rk} denotes a residual block that contains two $3\times3$ convolutional layers with the same number of filters on both layer. \texttt{uk} denotes a $3\times3$ fractional-strided-Convolution-InstanceNorm-ReLU layer with $k$ filters and stride $\frac{1}{2}$.

The network with 6 residual blocks consists of:\\
\texttt{c7s1-64,d128,d256,R256,R256,R256,\\
R256,R256,R256,u128,u64,c7s1-3}

The network with 9 residual blocks consists of:\\
\texttt{c7s1-64,d128,d256,R256,R256,R256,\\
R256,R256,R256,R256,R256,R256,u128} \\
\texttt{u64,c7s1-3}

{\bf Discriminator architectures}
For discriminator networks, we use $70\times 70$ PatchGAN~\cite{isola2016image}. 
Let \texttt{Ck} denote a $4\times4$ Convolution-InstanceNorm-LeakyReLU layer with k filters and stride $2$. After the last layer, we apply a convolution to produce a $1$-dimensional output. We do not use InstanceNorm for the first \texttt{C64} layer. We use leaky ReLUs with a slope of $0.2$. The discriminator architecture is:\\
\texttt{C64-C128-C256-C512}

\end{document}